%% file: sample-sigconf.tex
\documentclass[a4paper]{article}

\usepackage{booktabs} 

\usepackage{latexsym}
\usepackage{amssymb}
\usepackage{amsmath}
\usepackage{amsthm}
\usepackage{enumitem}
\usepackage{graphicx}
\usepackage{color,colortbl}

\usepackage{algorithm2e}

\usepackage{amsfonts}
\usepackage{pgfplots}
\pgfplotsset{compat=1.18}
\usepackage{lscape}
\usepackage{rotating}

\usepackage{float}

\usepackage{float}
\newtheorem{theorem}{Theorem}

\newcommand{\V}{\mathcal{V}}

\newcommand{\C}{\mathcal{K}}

\newcommand{\K}{{Cores}}

\usetikzlibrary{shapes.geometric}
\usetikzlibrary{positioning}

\usepackage{hyperref}

\title{Anytime Cooperative Implicit Hitting Set Solving}
\author{
  Emma Rollon \\
\and Javier Larrosa \\
\and Aleksandra Petrova \\
\multicolumn{1}{p{.8\textwidth}}{\centering\emph{Department of Computer Science, Universitat Politècnica de Catalunya, Spain}\\
}}

\begin{document}
\maketitle

\begin{abstract}
\noindent The \textit{Implicit Hitting Set} (HS) approach has shown to be very effective for MaxSAT, Pseudo-boolean optimization and other boolean frameworks. Very recently, it has also shown its potential in the very similar Weighted CSP framework by means of the so-called cost-function merging. The original formulation of the HS approach focuses on obtaining increasingly better lower bounds (HS-lb). However, and as shown for Pseudo-Boolean Optimization, this approach can also be adapted to compute increasingly better upper bounds (HS-ub). In this paper we consider both HS approaches and show how they can be easily combined in a multithread architecture where cores discovered by either component are available by the other which, interestingly, generates synergy between them. We show that the resulting algorithm (HS-lub) is consistently superior to either HS-lb and HS-ub in isolation. Most importantly, HS-lub has an effective anytime behaviour with which the optimality gap is reduced during the execution. We tested our approach on the Weighted CSP framework and show on three different benchmarks that our very simple implementation sometimes outperforms the parallel hybrid best-first search implementation of the far more developed state-of-the-art Toulbar2.
\end{abstract}

\section{Introduction}
Discrete Optimization problems are ubiquitous in life and solving them efficiently has attracted the interest of researchers for decades. When they are NP-complete, their optimization requires exponential time and sometimes is out of current technology. Then, anytime algorithms become crucial because they provide better and better solutions, the longer they keep running. An especially useful type of anytime algorithms are those that provide improving lower and upper bound of the optimum. They are very valuable because the optimality gap is an indication of the solution quality.

There exist several mathematical frameworks to model and solve discrete optimization problems. In this paper, we are concerned with \textit{Cost Function Networks}, which represent an additive objective function over many discrete variables. Cost Networks belong to a family of frameworks called \textit{Graphical Models} \cite{DBLP:conf/stacs/CooperGS20} which 
share the property of providing a concise description of multivariate functions using decomposability. While different queries can be made to a Cost Function Network, here we will consider the optimization of the sum of simple functions. This problem is often called the \textit{Weighted CSP} problem (WCSP) and it captures, among others, the \textit{Most Probable Explanation} in \textit{Markov Random Fields} and \textit{Bayesian Networks}. The WCSP problem has been found useful in a number of applications such as \textit{resource} \textit{allocation} \cite{DBLP:journals/constraints/CabonGLSW99}, \textit{bioinformatics} \cite{DBLP:journals/bioinformatics/ViricelGSB18,DBLP:journals/bioinformatics/VucinicSRBS20}, \textit{scheduling} \cite{DBLP:journals/constraints/BensanaLV99}, \textit{etc}. 

We focus on the recently proposed implicit hitting set (HS) approach for discrete optimization. It is a relatively generic solving paradigm that has been found successful in a variety of settings such as Max-SAT \cite{DBLP:phd/ca/Davies14,DBLP:conf/sat/BergBP20}, Pseudo-boolean Optimization (PBO) \cite{DBLP:conf/cp/IHS-PB1,DBLP:conf/sat/IHS-PB2} and Answer Set Programming (ASP) \cite{DBLP:conf/kr/SaikkoDAJ18}.
Regarding the WCSP problem, an HS based algorithm was proposed in \cite{DBLP:conf/cp/DelisleB13} and recently improved in \cite{Larrosa24}.

The HS approach is fundamentally an iterative process that implicitely contains a lower bound from a growing set of unsatisfiable pieces of the problem (a.k.a. cores). At each iteration, the current set of cores is deactivated (in HS terminology, hitted), which relaxes the original problem. The algorithm discovers new cores by solving the relaxation. The process ends when hitting all known cores yields a satisfiable relaxation. In its simplest form, the algorithm makes the implicit lower bound explicit at each iteration. Because of that, we will refer to this strategy as HS-lb.
Although, the strategy of HS-lb is essentially lower-bounding, many implementations also obtain sub-optimal solutions that are incidentally found during the process. The best-so-far of this solutions provides an upper bound of the optimum which can be used to stop the loop as soon as both bounds match. 

One of the bottlenecks in HS-lb is the computation of optimal hitting sets. An alternative is to compute cost-bounded hitting sets instead. This approach was suggested in~\cite{DBLP:conf/sat/IHS-PB2} and applied in the context of Pseudo-boolean Optimization. Our first contribution  is to adapt and test this idea to the context of WCSP. With this approach the emphasis is on the upper bound, and lower bound is not made explicit until the last iteration. Because of that, we will refer to it as HS-ub.



The second and main contribution of the paper is the description of an anytime algorithm in which both HS-lb and HS-ub are executed in parallel with a very simple shared-memory multithread implementation. In the resulting algorithm, called HS-lub, each algorithm computes improving bounds while collaborating with each other by sharing a common pool of cores and the bounds themselves. The interest of this approach is that cores found by one of the algorithms complement the cores found by the other. On one hand, cores found by HS-lb are more costly to obtain and more focused to the task of improving the lower bound. On the other hand, cores found by HS-ub are cheaper to obtain and more focused to the task of improving the upper bound. We empirically observe that they add some diversification to the set which, in turn, leads to an improved performance with respect to the isolated execution of both algorithms.


Although our approach can easily be adapted to Max-SAT, PBO and ASP, we focus on the Weighted CSPs framework. To evaluate its potential we compare its performance with the parallel hybrid best-first search of Toulbar2~\cite{DBLP:conf/cp/hbfs-multithread} on usual WCSP benchmarks. We show that our extremely simple implementation sometimes improves over the far more developed implementation of Toulbar2 which, we believe, shows the potential of our anytime HS proposal. 

\section{Related Work}

There are various algorithms that provide both lower and upper bounds. In the context of Max-SAT, so-called core-based algorithms solve instances by sequentially making calls to a SAT solver. From the sequence of calls to satisfiable formulas one can in general produce improving upper bounds. From the sequence of calls to unsatisfiable formulas one can extract cores which are aggregated by means of pseudo-boolean constraints producing improving lower bounds~\cite{DBLP:journals/ai/AnsoteguiBL13,DBLP:journals/constraints/MorgadoHLPM13}. In the general context of Graphical Models, we can distinguish between inference and search approaches. In the first case, we find the well-known mini-buckets-elimination (MBE) algorithm proposed in~\cite{minibuckets}. MBE has a parameter that trades time for accuracy. By iteratively increasing this parameter, the algorithm obtains a decreasing optimality gap. In the context of branch and bound systematic search some algorithms traverse the search space with hybrid strategies that combine best-first and depth-first. The best-first component provides a natural lower bound as the minimum cost among the heuristic value of all the open nodes. The depth-first component reaches near-optimal solutions which provide a natural upper bound. This idea has been long applied in Integer Programming solvers \cite{DBLP:books/daglib/nemhauser} and, more recently, in Weighted CSP \cite{DBLP:conf/cp/hbfs}.

Several parallel schemes have been proposed. An active area of research is the Distributed Constraint Reasoning~\cite{yeoh12}, which deals with complex synchronization strategies and a quite restricted environment. In the context of constraint solving (see~\cite{DBLP:journals/tplp/GentMNMPMU18} for a recent review), the concept of parallel tree search, where the search space is partitioned in some way into independent subproblems and each one is then solved in parallel, has been applied in~\cite{DBLP:journals/jair/MalapertRR16,DBLP:journals/jair/OttenD17,DBLP:conf/cp/hbfs}. In the context of Max-SAT, several solvers are also based in the parallel tree search and also on parallelized portfolio solvers~\cite{DBLP:books/sp/18/LynceMM18}.

\section{Preliminaries}
\subsection{CSPs and WCSPs}
A \textit{Constraint Satisfaction Problem} (CSP) is a tuple $(X,C)$ where $X$ is a set of \textit{variables} taking values in a finite domain, and $C$ is a set of \textit{constraints}. Each constraint  depends on a subset of variables called \textit{scope}. Constraints are boolean functions that forbid some of the possible assignments of the scope variables. A \textit{solution} is an assignment to every variable that satisfies all the constraints. Solving CSPs is an NP-complete problem \cite{DBLP:reference/fai/GentPP06}.

A \emph{Weighted CSP} (WCSP) is tuple $(X, C, F)$ where $(X, C)$ is a CSP and $F$ is a set of \emph{cost functions}.  A cost function $f\in F$ is a mapping that associates a cost to each possible assignment of the variables in its scope. The \emph{cost of a solution} is the sum of costs given by the different cost functions. The WCSP problem, which is known to be NP-hard \cite{DBLP:reference/fai/MeseguerRS06}, consists in computing a solution of minimum cost. 

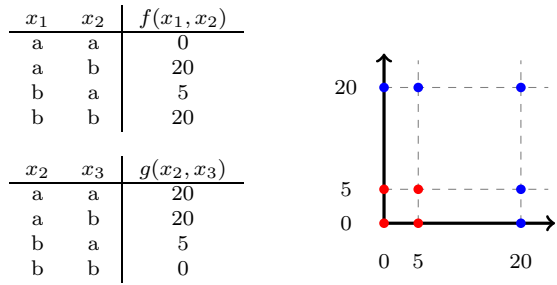
\begin{figure}[t]
\input{figuraEmma1}
    \caption{A WCSP with three variables $\{x_1,x_2,x_3\}$ and two cost-functions $F=\{f(\cdot),g(\cdot)\}$ (left). Its  vector space, where red and blue dots correspond to cores and solutions, respectively (right).}
    \label{fig:enter-label}
\end{figure}

Figure~\ref{fig:enter-label} (left) shows a WCSP with three variables $\{x_1, x_2, x_3\}$ having domain values $\{a,b\}$, no constraints and two cost functions $F=\{f(x_1,x_2),g(x_2,x_3)\}$. Because it does not have constraints, every assignment is a solution. An optimal solution is the assignment $x_1\leftarrow a, x_2\leftarrow b, x_3\leftarrow b$ with cost $20$.    

\subsection{Vectors and Dominance}
Given two vectors $\vec{u}$ and $\vec{v}$, the usual partial order among them, noted $\vec{u}\leq \vec{v}$, holds iff for each component $i$ we have that $u_i\leq v_i$. If $\vec{u}\leq \vec{v}$ we say that $\vec{v}$ \emph{dominates} $\vec{u}$. Given a set of vectors $\V$, we say that $\vec{u}$ \emph{hits} $\V$ if none of the vector in $\V$ dominates $\vec{u}$.



The \emph{minimum cost hitting vector} (MHV) of $\V$ is a vector that hits $\V$ with minimum cost. It is not difficult to see that MHV reduces to the classic \emph{hitting set problem}~\cite{DBLP:books/fm/GareyJ79} which, in its optimization version, is known to be NP-hard.

\subsection{Cores and Solutions}
In the following, we consider an arbitrary WCSP $(X,C,F)$ with $m$ cost functions $F=\{f_1,f_2,\ldots,f_m\}$. 
We name $w^*$ the cost of the optimal solution. 

A \textit{cost vector} $\vec{v}=(v_1,v_2,\ldots,v_m)$ is a vector where each component $v_i$ is associated to cost function $f_i$, and value $v_i$ must be a cost occurring in $f_i$. The \textit{cost} of vector $\vec{v}$ is $cost(\vec{v})=\sum_{i=1}^m v_i$. 

Vector $\vec{v}$ \textit{induces} a CSP $(X,C \cup F_{\vec{v}})$ where $F_{\vec{v}}$ denotes the set of constraints $f_i\leq v_i$, for $1\leq i \leq m$ (namely, cost functions are replaced by constraints). If the  CSP induced by $\vec{v}$ is satisfiable we say that $\vec{v}$ is a \textit{solution vector} (or simply a solution). Otherwise, we say that $\vec{v}$ is a \textit{core}. We denote the set of cores as $\K$. A core is \emph{maximal} if there is no other core in $\K$ that dominates it. An \textit{optimal solution} is a solution vector of minimum cost. It is easy to see that the cost of an optimal solution is $w^*$. 

Figure~\ref{fig:enter-label} (right) shows the space of vectors of our running WCSP example. Cores and solutions are represented by red and blue dots, respectively. Vector $\vec{v}=(5,5)$ is a core because, in CSP induced by $\vec{v}$, constraint $f(x_1,x_2)\leq 5$ is only satisfied by $x_2\leftarrow a$ and constraint $g(x_2,x_3)\leq 5$ is only satisfied by $x_2\leftarrow b$ (i.e., the two conditions are impossible to satisfy simultaneously). Note that there are $4$ cores but only vector $(5,5)$ is a maximal one (the other $3$ are dominated by $(5,5)$). One can easily see that vector $(20,0)$ is a solution vector because its induced CSP is satisfiable (the solution of the induced CSP being $x_1\leftarrow a, x_2\leftarrow b, x_3\leftarrow b$). Note as well that it is one of the two optimal solution vectors with cost $20$, the other one being $(0,20)$.

\section{Two HS-based Schemes}

The HS approach relies on the following, 

\begin{theorem}\label{obs1}
Consider a solution $\vec{h}$ and a set of cores $\C$. Then, $MHV(\C)\leq w^* \leq cost(\vec{h})$.
\end{theorem}

\noindent In words, the optimal cost $w^*$ is lower and upper bounded in terms of core and solution vectors.

In the following, we present two approaches that consider the previous Theorem from two different perspectives.

\subsection{HS-lb}
The HS approach to WCSP was first proposed in~\cite{DBLP:conf/cp/DelisleB13}, as a generalization of its application to MaxSAT~\cite{DBLP:phd/ca/Davies14}. 

Algorithm~\ref{alg:ihs1} (HS-lb) is a simple version of this idea. $\C$ is a set of cores, and $lb$ and $ub$ are the lower and upper bound of the optimum, respectively. 
At each itermation, the algorithm computes a minimum cost hitting vector $\vec{h}$. By Theorem~\ref{obs1}, we know that the cost of $\vec{h}$ is a lower bound of the problem, so the algorithm updates $lb$. Then, it solves the CSP induced by $\vec{h}$. If it is satisfiable (i.e., $\vec{h}$ is a solution), then we know that $lb=ub$ so the algorithm can stop. Otherwise, vector $\vec{h}$ is a core, and the algorithm computes a new maximal core $\vec{k}$ such that $\vec{h}\leq \vec{k}$ and adds it to  $\C$.


\begin{algorithm}[t]
  \caption{Baseline HS algorithm for WCSP. It receives a WCSP $(X,C,F)$ and returns its optimal cost $w^*$. Function $MaximalCore(\cdot)$ receives a core $\vec{h}$ and returns a maximal dominating core $\vec{k}$.}
  \label{alg:ihs1}
 
  \SetKwFunction{SolveWCSP}{HS-lb}
  \SetKwFunction{Solve}{SolveCSP}
  \SetKwFunction{SolveMin}{MinCostHittingVector}
  \SetKwFunction{Grow}{MaximalCore}
  {\bf Function} \SolveWCSP{$X,C,F$}\\
  \Begin{
    $\C:=\emptyset;\ lb:=0;\ ub:=\infty\ $\;
    \While{$lb<ub$}{
$\vec{h}:=$\SolveMin{$\C$}\;
$lb:=cost(\vec{h})$\;
\lIf{\Solve{$X,C \cup F_{\vec{h}}$}}{
    $\ ub:=cost(\vec{h})$}
    \Else{
     $\vec{k}:=$\Grow{$X,C,F,ub,\vec{h}$}\;
     $\C:=\C \cup \{\vec{k}\}$\;     
    }
}
\Return $lb$
}

\end{algorithm}

\begin{algorithm}[t]
  \LinesNumbered
  \caption{Alternative HS approach for WCSP.}
  \label{alg:ihsub}
 
  \SetKwFunction{SolveWCSP}{HS-ub}
  \SetKwFunction{Solve}{SolveCSP}
  \SetKwFunction{SolveMin}{CostBoundedHV}
  \SetKwFunction{Grow}{MaximalCore}
  {\bf Function} \SolveWCSP{$X,C,F$}\\
  \Begin{
    $\C:=\emptyset;\ lb:=0;\ ub:=\infty\ $\;
    \While{$lb<ub$}{
$\vec{h}:=$\SolveMin{$\C,ub$}\;
\lIf{$\vec{h}=$NUL}{$lb:=ub$}
\Else{
\lIf{\Solve{$X,C \cup F_{\vec{h}}$}}{$ub:=cost(\vec{h})$}
\Else{
     $\vec{k}:=$\Grow{$X,C,F,ub,\vec{h}$}\;
     $\C:=\C \cup \{\vec{k}\}$\;     
    }
    }
}
\Return $lb$
}

\end{algorithm}

\subsection{HS-ub}

Each iteratin of HS-lb may be very time consuming because it needs to compute minimum cost hitting vector $\vec{h}$ which is an NP-hard problem.
One way to decrease the work-load of each iteration is to rely on non-optimal hitting vectors. As suggested in~\cite{DBLP:conf/sat/IHS-PB2}, we can replace optimal hitting vectors by hitting vectors of bounded cost. 

Algorithm~\ref{alg:ihsub} (HS-ub) implements this idea. As before, $\C$ is a set of cores, and $lb$ and $ub$ are the lower and upper bound of the optimum. At each iteration, the algorithm computes a hitting vector $\vec{h}$ with cost less than $ub$. If such $\vec{h}$ does not exists (i.e., it is NUL), it means that, by Theorem~\ref{obs1}, the current ub is the optimum so the algorithm can stop. If $\vec{h}$ exists, then the CSP induced by $\vec{h}$ is solved. If it is satisfiable (i.e., $\vec{h}$ is a solution) the upper bound is updated. Otherwise, vector $\vec{h}$ is a core, and the algorithm computes a new maximal core $\vec{k}$ which is added to $\C$.

\subsection{Computing Maximal Cores}
It is important to note that both algorithms are correct even if $\vec{h}$ is directly added to $\C$ (i.e., skipping the $MaximalCore(\cdot)$ call). However, as suggested in~\cite{DBLP:phd/ca/Davies14} and supported by our own experiments, adding a larger core is fundamental for both HS-lb and HS-ub. The rationale is that adding maximal cores to $\C$ results in larger (i.e., with higher cost) hitting vectors. 
As a consequence, when $\C$ is hit optimally, the lower bound grows faster, while when $\C$ is hit non-optimally, the area under the $ub$ is further reduced.


The natural implementation of the $MaximalCore(\cdot)$ function is to obtain $\vec{k}$ as a sequence of increments to its components while preserving the core property until no more increments can be done. After each increment the algorithm needs to check if the new induced CSP remains unsatisfiable (i.e., a call to $SolveCSP(\cdot)$). Interestingly, during this process solution vectors are found, which means that the upper bound may be lowered inside the $MaximalCore(\cdot)$ function.

\subsection{Discussion}

\begin{figure}[t]
    \input{figuraEmma2}
    \caption{Graphical representation of an arbitrary iteration of HS-lb ($\vec{h}$ and $\vec{k}$) and HS-ub ($\vec{h}'$ and $\vec{k}$). Current $lb$ is $6$, the optimal value has cost $w^*=8$ and $ub>10$. Blue dots represent solutions, red dots cores and the green dot the optimal solution.}
    \label{fig:it}
\end{figure}

The main advantage of HS-ub compared to HS-lb is that iterations are likely to be faster. There are several reasons for it. First, it is much more efficient to find a bounded hitting vector (which is a decision problem) than finding an optimal hitting vector (which is an optimization problem). Second, only in the last call of $CostBoundedHS(\cdot)$ the problem will be unsatisfiable (which is typically a much more costly task to solve). Finally, the returned cost-bounded hitting vectors are likely to have a higher cost (near $ub$) because they are easier to find. Therefore, obtaining a maximal core $\vec{k}$ will not need so many $SolveCSP(\cdot)$ calls.

The advantage is at the cost of potentially more iterations. On the one hand, in HS-ub not all iterations end up adding a new core because some iterations, when the induced CSP is satisfiable, only decrease the upper bound. On the other hand, the cost bounded hitting vectors may not contribute to increase the domination of the area of interest either. This idea is graphically depicted in Figure~\ref{fig:it}. HS-lb would compute (the only) optimal hitting vector $\vec{h}$ and would transform it into a maximal core $\vec{k}$. However, there are many hitting vectors with cost less than $ub$ that HS-ub could compute and one possibility would be $\vec{h}'$ and a possible maximal core could be $\vec{k}'$. Note that in this example, the iteration of HS-ub albeit cheaper, is completely wasted because adding $\vec{k}'$ to the set of cores does not contribute to the domination of minimum cost cores. 

\section{Any-time HS scheme}

\begin{figure}[t]
    \centering
\scalebox{0.5}{
\begin{tikzpicture}
\begin{axis}[
    title={Spot 42},
    ylabel={cost},
    xmax=90,
    ymax=160000,    
    ymin=150000,
    legend pos=north east,
    legend cell align=left,
    label style={font=\LARGE},
    tick label style={font=\Large},
    title style={font=\LARGE}
]

\addplot [blue] table [header=false, x expr={\thisrowno{1}}, y expr={(\thisrowno{0}==-1?\thisrowno{2}:NaN)}] {./dat/42_VAC_ihs_merged_a1_t4.dat}; 
\addplot  [blue] table [header=false, x expr={\thisrowno{1}}, y expr={(\thisrowno{0}==-2?\thisrowno{2}:NaN)}] {./dat/42_VAC_ihs_merged_a1_t4.dat}; 
\addplot  [red, dashed] table [header=false, x expr={\thisrowno{0}}, y expr={\thisrowno{1}}] {./dat/42_VAC_ihs_merged_a2_t4.dat}; 
\end{axis}
\end{tikzpicture}
\hspace{0.5cm}
\begin{tikzpicture}
\begin{axis}[
    title={pedigree41},
    ymax=8000,    
    ymin=6000,
    legend pos=north east,
    legend cell align=left,
    label style={font=\LARGE},
    tick label style={font=\Large, /pgf/number format/.cd, 1000 sep={}},
    title style={font=\LARGE},
    legend style={font=\LARGE}
]

\addplot [blue] table [header=false, x expr={\thisrowno{1}}, y expr={(\thisrowno{0}==-1?\thisrowno{2}:NaN)}] {./dat/pedigree41_VAC_ihs_merged_a1_t4.dat}; 
\addplot  [blue] table [header=false, x expr={\thisrowno{1}}, y expr={(\thisrowno{0}==-2?\thisrowno{2}:NaN)}] {./dat/pedigree41_VAC_ihs_merged_a1_t4.dat}; 
\addplot  [red, dashed] table [header=false, x expr={\thisrowno{0}}, y expr={\thisrowno{1}}] {./dat/pedigree41_VAC_ihs_merged_a2_t4.dat}; 

  \legend{HS-lb, , HS-ub}
\end{axis}
\end{tikzpicture}
}


\scalebox{0.5}{
\begin{tikzpicture}
\begin{axis}[
    xlabel={time (seconds)},
    ylabel={nb cores},
    legend pos=north east,
    legend cell align=left,
    label style={font=\Large},
    tick label style={font=\Large},
    title style={font=\Large}
]

\addplot [blue] table [header=false, x index=0, y index=1] {./dat/42_VAC.wcsp_ihs_merged_a1_t4.cores}; 
\addplot  [red, dashed] table [header=false, x index=0, y index=1] {./dat/42_VAC.wcsp_ihs_merged_a2_t4.cores}; 
\end{axis}
\end{tikzpicture}
\hspace{0.7cm}
\begin{tikzpicture}
\begin{axis}[
    xlabel={time (seconds)},
    legend pos=north east,
    legend cell align=left,
    label style={font=\Large},
    tick label style={font=\Large},
    title style={font=\Large}
]

\addplot [blue] table [header=false, x index=0, y index=1] {./dat/pedigree41_VAC.wcsp_ihs_merged_a1_t4.cores}; 
\addplot  [red, dashed] table [header=false, x index=0, y index=1] {./dat/pedigree41_VAC.wcsp_ihs_merged_a2_t4.cores}; 

\end{axis}
\end{tikzpicture}
}
\caption{Two selected executions of HS-lb (solid blue) and HS-ub (dashed red). Plots on the top row (resp. bottom row) show the evolution of bounds (resp. number of accumulated cores) as a function of time. }
\label{fig:evol}
\end{figure}
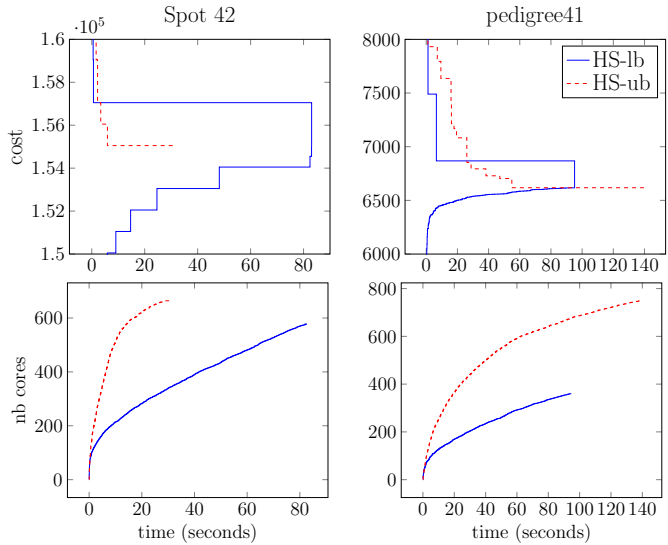

We discussed in the previous section that HS-lb and HS-ub are complementary implementations aiming at the same goal: finding a set of cores that dominates every core with cost less than the optimum. HS-lb is likely to be slower but its cores are likely to be of \textit{better quality}. HS-ub is likely to be faster but it is more likely to need more cores. Most importantly, none of them is guaranteed to outperform the other. Figure~\ref{fig:evol} nicely illustrates our discussion by means of two selected instances from standard benchmarks. Plots on the left correspond to a scheduling instance, plots on the right correspond to a bioinformatics instance. Plots on the top show the evolution of the bounds for both HS-lb and HS-ub as a function of time. Only HS-lb offers both bounds. It can be seen that HS-ub provides in both cases a better upper bound. Note that in the first instance HS-lb outperforms HS-ub and in the second instance it is the other way around. The plots on the bottom show the number of cores that both algorithms accumulate along time. We can see that in both cases HS-ub adds cores much faster (namely, iterates faster), but HS-lb requires less cores to achieve the termination condition (namely, computes more useful cores).

Since both algorithms consider the core extraction task with complementary strengths, a natural question arises: would they complement each other if they collaborate in the search for new cores?  Our hypothesis is that, by sharing the set of cores $\C$ and the bounds ($lb$ and $ub$) found so far, the cheaper cores computed by HS-ub will help HS-lb to find better lower bounds which, in turn, will help HS-ub to improve the upper bounds. 

The parallel algorithm that implements this idea, called HS-lub, is a straightforward shared memory implementation where both HS-lb and HS-ub run in parallel. Each process uses mutual exclusion to lock the shared memory locations and obtain exclusive access when they need to update them.








\section{Experimental Results}

\subsection{Evaluation setting}

We implemented all our algorithms in C++. The \emph{SolveCSP} function encodes the induced CSP $F_{\vec{h}}$ as a CNF formula and uses the assumption-based SAT solver \textit{CaDiCal}~\cite{cadical} to solve it. Both, \emph{MinCostHittingVector} and \emph{CostBoundedHV} are implemented as a 0-1 integer program solved with CPLEX \cite{cplex2009v12}. Our encodings are similar to~\cite{DBLP:conf/cp/DelisleB13}). For the parallel HS-lub, we use POSIX threads to implement the parallelism.

We enhance the basic description of the algorithms with cost-function merging~\cite{Larrosa24} and disjoint cores~\cite{DBLP:conf/cp/DelisleB13}. There are a number of additional improvements that we did not incorporate such as reduced cost fixing~\cite{DBLP:conf/cp/IHS-PB1}, weight-aware cost extraction \cite{DBLP:conf/sat/IHS-PB2} or greedy hitting vectors~\cite{DBLP:phd/ca/Davies14,DBLP:conf/cp/DelisleB13}. Given the good results of using these techniques in Max-SAT and PBO, we expect that their incorporation would have boosted all HS algorithms even further.

We perform our empirical evaluation over four benchmarks selected from~\cite{DBLP:journals/constraints/HurleyOAKSZG16} with a total of $304$ instances: SPOT5 (satellite scheduling) with $20$ instances, Pedigree (linkage analysis) with $22$ instances, Maxclique (DIMACS maximum clique) with $62$ instances, and EHI (Random 3-SAT instances embedding a small unsatisfiable part and converted into a binary CSP) with $200$ instances. Note that Pedigree and Maxclique instances were also considered in the empirical evaluation of parallel hbfs in~\cite{DBLP:conf/cp/hbfs-multithread}, which is the current state-of-the-art. 

We run all the experiments on a single multi-core machine running Linux with $2.20$ GHz CPU, $128$GB RAM memory and $16$ cpu-cores. Solving times are reported in seconds and correspond to CPU (resp. wall-clock) time for the sequential (resp. parallel) methods. The time limit is $30$ minutes. We discarded instances that all algorithms solved in less than a second ($2$ instances in SPOT5, $4$ instances in Pedigree, $5$ instances in Maxclique and $0$ in EHI). 

Note that CaDiCal is not multithread, but CPLEX is. Therefore, HS-lb and HS-ub use $16$ cpu-cores during the computation of the hitting vector and $1$ cpu-core during its improvement. HS-lub has two instances of CPLEX running in parallel and we can choose how many cpu-cores are given to each one. 

We run parallel hbfs on the $16$ cpu-cores and as described in~\cite{DBLP:conf/cp/hbfs-multithread} (i.e., using default options in Toulbar2 v-1.2 except no dichotomic branching (option -d:), and using the burst mode).

\subsection{Results}

\begin{landscape}
\begin{table*}[th]
\scriptsize
\begin{tabular}{l|rrrrr|rrrr|rrr|rrrr}
\hline
 & \multicolumn{5}{c|}{HS-lub} & \multicolumn{4}{c|}{HS-lb} &  \multicolumn{3}{c|}{HS-ub} &  \multicolumn{4}{c}{hbfs-16} \\  
Inst. & lb & ub &  gap  &cores& t & lb & ub & cores & t & ub & cores & t & lb & ub & gap & t \\ \hline
1401 & {\cellcolor{gray!30}\bf{438606}} & 469101 & {\cellcolor{gray!30}10.25} & 10747 & - & 431104 & 490109 & \bf{810} & - & \bf{469099} & 8697 & - & 183552 & {\cellcolor{gray!30}465104} & 95.91 & - \\ 
 1403 & {\cellcolor{gray!30}\bf{437745}} & {\cellcolor{gray!30}466255} & {\cellcolor{gray!30}9.68} & 10346 & - & 431244 & 486257 & \bf{1133} & - & \bf{466254} & 9934 & - & 183605 & 470256 & 95.98 & - \\ 
 1405 & {\cellcolor{gray!30}\bf{436902}} & 471424 & {\cellcolor{gray!30}11.52} & 12315 & - & 430397 & 492434 & \bf{1713} & - & \bf{469417} & 14120 & - & 180644 & {\cellcolor{gray!30}463424} & 96.92 & - \\ 
 1407 & {\cellcolor{gray!30}\bf{438549}} & {\cellcolor{gray!30}\bf{468583}} & {\cellcolor{gray!30}10.12} & 14669 & - & 429542 & 494612 & \bf{2096} & - & 468589 & 13365 & - & 180674 & 469622 & 96.98 & - \\ 
 1506 & {\cellcolor{gray!30}\bf{350507}} & {\cellcolor{gray!30}\bf{355537}} & {\cellcolor{gray!30}2.38} & 11560 & - & 346005 & 370539 & \bf{2336} & - & 357527 & 13479 & - & 153174 & 358534 & 95.80 & - \\ 
 28 & {\cellcolor{gray!30}\bf{255103}} & \bf{272105} & {\cellcolor{gray!30}10.12} & 9652 & - & 250607 & 280105 & \bf{1256} & - & \bf{272105} & 8973 & - & 180562 & {\cellcolor{gray!30}270105} & 53.93 & - \\ 
 5 & {\cellcolor{gray!30}\bf{245}} & {\cellcolor{gray!30}\bf{263}} & {\cellcolor{gray!30}8.96} & 19974 & - & 234 & 271 & \bf{845} & - & 264 & 17549 & - & 71 & 270 & 95.67 & - \\  
  1504 & {\cellcolor{gray!30}\bf{161287}} & {\cellcolor{gray!30}\bf{161287}} & {\cellcolor{gray!30}0} & 3289 & {\cellcolor{gray!30}\bf{50.22}} & \bf{161287} & \bf{161287} & \bf{1207} & 151.49 & \bf{161287} & 2014 & - & 104624 & 161289 & 91.89 & - \\ 
 408 & {\cellcolor{gray!30}\bf{6228}} & {\cellcolor{gray!30}\bf{6228}} & {\cellcolor{gray!30}0} & 320 & {\cellcolor{gray!30}\bf{1.69}} & \bf{6228} & \bf{6228} & \bf{220} & 7.11 & \bf{6228} & 288 & 3.75 & 4186 & {\cellcolor{gray!30}6228} & 96.96 & - \\ 
 412 & {\cellcolor{gray!30}\bf{32381}} & {\cellcolor{gray!30}\bf{32381}} & {\cellcolor{gray!30}0} & 580 & {\cellcolor{gray!30}\bf{7.13}} & \bf{32381} & \bf{32381} & \bf{374} & 19.71 & \bf{32381} & 621 & 16.38 & 23706 & {\cellcolor{gray!30}32381} & 56.99 & - \\ 
 414 & {\cellcolor{gray!30}\bf{38478}} & {\cellcolor{gray!30}\bf{38478}} & {\cellcolor{gray!30}0} & 526 & {\cellcolor{gray!30}\bf{36.20}} & \bf{38478} & \bf{38478} & \bf{410} & 88.19 & \bf{38478} & 463 & 98.93 & 23675 & 38479 & 80.91 & - \\ 
 42 & {\cellcolor{gray!30}\bf{155050}} & {\cellcolor{gray!30}\bf{155050}} & {\cellcolor{gray!30}0} & 707 & {\cellcolor{gray!30}\bf{20.59}} & \bf{155050} & \bf{155050} & \bf{578} & 83.06 & \bf{155050} & 664 & 31.64 & 109045 & {\cellcolor{gray!30}155050} & 55.76 & - \\ 
 505 & {\cellcolor{gray!30}\bf{21253}} & {\cellcolor{gray!30}\bf{21253}} & {\cellcolor{gray!30}0} & 324 & {\cellcolor{gray!30}\bf{2.17}} & \bf{21253} & \bf{21253} & \bf{201} & 5.65 & \bf{21253} & 238 & 3.92 & 15133 & {\cellcolor{gray!30}21253} & 66.98 & - \\ 
 507 & {\cellcolor{gray!30}\bf{27390}} & {\cellcolor{gray!30}\bf{27390}} & {\cellcolor{gray!30}0} & 430 & {\cellcolor{gray!30}\bf{6.95}} & \bf{27390} & \bf{27390} & \bf{298} & 18.76 & \bf{27390} & 429 & 27.60 & 19667 & {\cellcolor{gray!30}27390} & 63.17 & - \\ 
 509 & {\cellcolor{gray!30}\bf{36446}} & {\cellcolor{gray!30}\bf{36446}} & {\cellcolor{gray!30}0} & 564 & {\cellcolor{gray!30}\bf{24.77}} & \bf{36446} & \bf{36446} & \bf{406} & 51.20 & \bf{36446} & 553 & 58.88 & 30686 & {\cellcolor{gray!30}36446} & 33.35 & - \\ 
 1502 & {\cellcolor{gray!30}\bf{28042}} & {\cellcolor{gray!30}\bf{28042}} & {\cellcolor{gray!30}0} & 3 & {\cellcolor{gray!30}\bf{0.02}} & \bf{28042} & \bf{28042} & \bf{2} & \bf{0.02} & \bf{28042} & \bf{2} & 0.03 & {\cellcolor{gray!30}28042} & {\cellcolor{gray!30}28042} & {\cellcolor{gray!30}0} & 0.05 \\ 
 404 & {\cellcolor{gray!30}\bf{114}} & {\cellcolor{gray!30}\bf{114}} & {\cellcolor{gray!30}0} & 108 & {\cellcolor{gray!30}\bf{0.24}} & \bf{114} & \bf{114} & \bf{65} & 0.54 & \bf{114} & 84 & 0.59 & {\cellcolor{gray!30}114} & {\cellcolor{gray!30}114} & {\cellcolor{gray!30}0} & 5.43 \\ 
 503 & {\cellcolor{gray!30}\bf{11113}} & {\cellcolor{gray!30}\bf{11113}} & {\cellcolor{gray!30}0} & 78 & {\cellcolor{gray!30}\bf{0.11}} & \bf{11113} & \bf{11113} & 42 & 0.21 & \bf{11113} & \bf{36} & 0.17 & {\cellcolor{gray!30}11113} & {\cellcolor{gray!30}11113} & {\cellcolor{gray!30}0} & 605.31 \\ 
 \hline
 ped40 & {\cellcolor{gray!30}\bf{7226}} & {\cellcolor{gray!30}7334} & {\cellcolor{gray!30}2.97} & 5486 & - & 7182 & 7982 & \bf{1247} & - & \bf{7330} & 6658 & - & 4526 & 7454 & 77.98 & - \\  
 ped19 & {\cellcolor{gray!30}\bf{4625}} & {\cellcolor{gray!30}\bf{4625}} & {\cellcolor{gray!30}0} & 2840 & {\cellcolor{gray!30}\bf{526.02}} & \bf{4625} & \bf{4625} & \bf{1055} & 899.81 & \bf{4625} & 3545 & 707.65 & 2652 & 4730 & 74.99 & - \\  
 ped51 & {\cellcolor{gray!30}\bf{6406}} & {\cellcolor{gray!30}\bf{6406}} & {\cellcolor{gray!30}0} & 234 & {\cellcolor{gray!30}\bf{2.37}} & \bf{6406} & \bf{6406} & \bf{171} & 6.42 & \bf{6406} & 570 & 16.04 & 5523 & {\cellcolor{gray!30}6406} & 35.45 & - \\  
 ped13 & {\cellcolor{gray!30}\bf{2030}} & {\cellcolor{gray!30}\bf{2030}} & {\cellcolor{gray!30}0} & 197 & \bf{3.41} & \bf{2030} & \bf{2030} & \bf{159} & 9.89 & \bf{2030} & 201 & 7.25 & {\cellcolor{gray!30}2030} & {\cellcolor{gray!30}2030} & {\cellcolor{gray!30}0} & {\cellcolor{gray!30}1.79} \\ 
 ped18 & {\cellcolor{gray!30}\bf{7021}} & {\cellcolor{gray!30}\bf{7021}} & {\cellcolor{gray!30}0} & 215 & {\cellcolor{gray!30}\bf{2.75}} & \bf{7021} & \bf{7021} & \bf{117} & 4.27 & \bf{7021} & 398 & 19.97 & {\cellcolor{gray!30}7021} & {\cellcolor{gray!30}7021} & {\cellcolor{gray!30}0} & 10.79 \\ 
 ped20 & {\cellcolor{gray!30}\bf{2532}} & {\cellcolor{gray!30}\bf{2532}} & {\cellcolor{gray!30}0} & 94 & \bf{0.91} & \bf{2532} & \bf{2532} & \bf{69} & 2.43 & \bf{2532} & 91 & 1.65 & {\cellcolor{gray!30}2532} & {\cellcolor{gray!30}2532} & {\cellcolor{gray!30}0} & {\cellcolor{gray!30}0.19} \\ 
 ped23 & {\cellcolor{gray!30}\bf{2489}} & {\cellcolor{gray!30}\bf{2489}} & {\cellcolor{gray!30}0} & 62 & \bf{0.50} & \bf{2489} & \bf{2489} & \bf{36} & 1.01 & \bf{2489} & 42 & 0.91 & {\cellcolor{gray!30}2489} & {\cellcolor{gray!30}2489} & {\cellcolor{gray!30}0} & {\cellcolor{gray!30}0.04} \\ 
 ped25 & {\cellcolor{gray!30}\bf{10630}} & {\cellcolor{gray!30}\bf{10630}} & {\cellcolor{gray!30}0} & 148 & {\cellcolor{gray!30}\bf{1.47}} & \bf{10630} & \bf{10630} & \bf{89} & 3.23 & \bf{10630} & 170 & 6.24 & {\cellcolor{gray!30}10630} & {\cellcolor{gray!30}10630} & {\cellcolor{gray!30}0} & 529.77 \\ 
 ped30 & {\cellcolor{gray!30}\bf{7341}} & {\cellcolor{gray!30}\bf{7341}} & {\cellcolor{gray!30}0} & 230 & {\cellcolor{gray!30}\bf{2.80}} & \bf{7341} & \bf{7341} & \bf{125} & 4.22 & \bf{7341} & 585 & 31.66 & {\cellcolor{gray!30}7341} & {\cellcolor{gray!30}7341} & {\cellcolor{gray!30}0} & 45.88 \\ 
 ped31 & {\cellcolor{gray!30}\bf{5258}} & {\cellcolor{gray!30}\bf{5258}} & {\cellcolor{gray!30}0} & 2124 & {\cellcolor{gray!30}\bf{377.58}} & \bf{5258} & \bf{5258} & \bf{1016} & 715.06 & \bf{5258} & 2816 & 711.53 & {\cellcolor{gray!30}5258} & {\cellcolor{gray!30}5258} & {\cellcolor{gray!30}0} & 399.93 \\ 
 ped33 & {\cellcolor{gray!30}\bf{5855}} & {\cellcolor{gray!30}\bf{5855}} & {\cellcolor{gray!30}0} & 115 & {\cellcolor{gray!30}\bf{0.59}} & \bf{5855} & \bf{5855} & \bf{76} & 1.55 & \bf{5855} & 82 & 1.50 & {\cellcolor{gray!30}5855} & {\cellcolor{gray!30}5855} & {\cellcolor{gray!30}0} & 0.72 \\ 
 ped34 & {\cellcolor{gray!30}\bf{6174}} & {\cellcolor{gray!30}\bf{6174}} & {\cellcolor{gray!30}0} & 157 & \bf{2.92} & \bf{6174} & \bf{6174} & \bf{93} & 5.05 & \bf{6174} & 213 & 10.24 & {\cellcolor{gray!30}6174} & {\cellcolor{gray!30}6174} & {\cellcolor{gray!30}0} & {\cellcolor{gray!30}0.51} \\ 
 ped37 & {\cellcolor{gray!30}\bf{9080}} & {\cellcolor{gray!30}\bf{9080}} & {\cellcolor{gray!30}0} & 188 & \bf{1.03} & \bf{9080} & \bf{9080} & \bf{113} & 1.92 & \bf{9080} & 253 & 5.43 & {\cellcolor{gray!30}9080} & {\cellcolor{gray!30}9080} & {\cellcolor{gray!30}0} & {\cellcolor{gray!30}0.53} \\ 
 ped39 & {\cellcolor{gray!30}\bf{11793}} & {\cellcolor{gray!30}\bf{11793}} & {\cellcolor{gray!30}0} & 281 & \bf{3.81} & \bf{11793} & \bf{11793} & \bf{232} & 13.35 & \bf{11793} & 309 & 7.91 & {\cellcolor{gray!30}11793} & {\cellcolor{gray!30}11793} & {\cellcolor{gray!30}0} & {\cellcolor{gray!30}3.51} \\ 
 ped41 & {\cellcolor{gray!30}\bf{6618}} & {\cellcolor{gray!30}\bf{6618}} & {\cellcolor{gray!30}0} & 735 & \bf{81.71} & \bf{6618} & \bf{6618} & \bf{361} & 95.16 & \bf{6618} & 749 & 139.63 & {\cellcolor{gray!30}6618} & {\cellcolor{gray!30}6618} & {\cellcolor{gray!30}0} & {\cellcolor{gray!30}42.40} \\ 
 ped44 & {\cellcolor{gray!30}\bf{6651}} & {\cellcolor{gray!30}\bf{6651}} & {\cellcolor{gray!30}0} & 457 & {\cellcolor{gray!30}\bf{14.12}} & \bf{6651} & \bf{6651} & \bf{252} & 23.51 & \bf{6651} & 616 & 34.54 & {\cellcolor{gray!30}6651} & {\cellcolor{gray!30}6651} & {\cellcolor{gray!30}0} & 1119.91 \\ 
 ped7 & {\cellcolor{gray!30}\bf{3548}} & {\cellcolor{gray!30}\bf{3548}} & {\cellcolor{gray!30}0} & 71 & {\cellcolor{gray!30}\bf{0.30}} & \bf{3548} & \bf{3548} & \bf{45} & 0.51 & \bf{3548} & 102 & 1.57 & {\cellcolor{gray!30}3548} & {\cellcolor{gray!30}3548} & {\cellcolor{gray!30}0} & 1.78 \\ 
 ped9 & {\cellcolor{gray!30}\bf{7040}} & {\cellcolor{gray!30}\bf{7040}} & {\cellcolor{gray!30}0} & 890 & {\cellcolor{gray!30}\bf{41.16}} & \bf{7040} & \bf{7040} & \bf{452} & 93.24 & \bf{7040} & 1312 & 81.67 & {\cellcolor{gray!30}7040} & {\cellcolor{gray!30}7040} & {\cellcolor{gray!30}0} & 872.58 \\ 
  \hline

EHI-85 & {\cellcolor{gray!30}\bf{9}} & {\cellcolor{gray!30}\bf{9}} & {\cellcolor{gray!30}0} & 17.28 & {\cellcolor{gray!30}\bf{18.5}} & \bf{9} & \bf{9} & 13.20 & 22.2 & \bf{9} & \bf{11.93} & 20.4 & {\cellcolor{gray!30}9} & {\cellcolor{gray!30}9} & {\cellcolor{gray!30}0} & 53.4 \\ 
 EHI-90 & {\cellcolor{gray!30}\bf{9}} & {\cellcolor{gray!30}\bf{9}} & {\cellcolor{gray!30}0} & 18.01 & {\cellcolor{gray!30}\bf{19.9}} & \bf{9} & \bf{9} & 13.49 & 22.1 & \bf{9} & \bf{12.12} & 22.0 & {\cellcolor{gray!30}9} & {\cellcolor{gray!30}9} & {\cellcolor{gray!30}0} & 74.7 \\ 
\hline
\end{tabular}
\caption{Comparison of HS-lub, HS-lb, HS-ub and hbfs on SPOT, Pedigree and EHI instnaces. All algorithms are executed with 16 cpu-cores. Bodface is used to emphasize winning values when comparing HS-lub with respect to HS-lb and HS-ub. Grey back-ground is used to emphasize winning values when comparing HS-lub with respect to hbfs.
An "-" indicates time-limit exceeded. For EHI-85 and EHI-90, we only report mean values over their 100 instances, respectively.}\label{table-benchs}
\end{table*}
\end{landscape}

\begin{landscape}
\begin{table*}[th]
\scriptsize
\begin{tabular}{l|rrrrr|rrrr|rrr|rrrr}
\hline
 & \multicolumn{5}{c|}{HS-lub} & \multicolumn{4}{c|}{HS-lb} &  \multicolumn{3}{c|}{HS-ub} &  \multicolumn{4}{c}{hbfs-16} \\  
Inst. & lb & ub &  gap  &cores& t & lb & ub & cores & t & ub & cores & t & lb & ub & gap & t \\ \hline
 MANN-a81 & {\cellcolor{gray!30}2191} & {\cellcolor{gray!30}\bf{2221}} & {\cellcolor{gray!30}2.8} & 5309 & - & \bf{2204} & 2372 & \bf{2867} & - & \bf{2221} & 5617 & - & 1999 & {\cellcolor{gray!30}2221} & 20.9 & - \\ 
 brock400-1 & {\cellcolor{gray!30}\bf{338}} & \bf{379} & {\cellcolor{gray!30}22.9} & 2080 & - & 335 & 380 & \bf{887} & - & \bf{379} & 1240 & - & 327 & {\cellcolor{gray!30}376} & 27.8 & - \\ 
 brock400-2 & {\cellcolor{gray!30}\bf{337}} & 380 & 23.9 & 1932 & - & 335 & 381 & \bf{1020} & - & \bf{379} & 1793 & - & 333 & {\cellcolor{gray!30}372} & {\cellcolor{gray!30}22.7} & - \\ 
 brock400-3 & {\cellcolor{gray!30}\bf{332}} & \bf{379} & 26.3 & 1054 & - & 323 & \bf{379} & \bf{200} & - & \bf{379} & 358 & - & 331 & {\cellcolor{gray!30}369} & {\cellcolor{gray!30}22.5} & - \\ 
 brock400-4 & {\cellcolor{gray!30}\bf{336}} & \bf{379} & 24.0 & 2578 & - & 332 & 381 & \bf{947} & - & 380 & 2618 & - & 334 & {\cellcolor{gray!30}367} & {\cellcolor{gray!30}19.8} & - \\ 
 brock800-1 & {\cellcolor{gray!30}\bf{682}} & 785 & {\cellcolor{gray!30}26.8} & 51 & - & \bf{682} & 785 & 35 & - & \bf{783} & \bf{23} & - & 640 & {\cellcolor{gray!30}780} & 36.8 & - \\ 
 brock800-2 & \bf{400} & \bf{800} & 100.0 & 0 & - & \bf{400} & \bf{800} & 0 & - & \bf{800} & 0 & - & {\cellcolor{gray!30}628} & {\cellcolor{gray!30}779} & {\cellcolor{gray!30}39.8} & - \\ 
 brock800-3 & {\cellcolor{gray!30}691} & \bf{784} & {\cellcolor{gray!30}24.2} & 103 & - & \bf{695} & \bf{784} & 91 & - & 787 & \bf{19} & - & 632 & {\cellcolor{gray!30}779} & 38.8 & - \\ 
 brock800-4 & {\cellcolor{gray!30}\bf{667}} & \bf{784} & {\cellcolor{gray!30}30.5} & 61 & - & \bf{667} & \bf{784} & 41 & - & 786 & \bf{20} & - & 651 & {\cellcolor{gray!30}780} & 33.9 & - \\ 
 hamming10-4 & {\cellcolor{gray!30}\bf{887}} & \bf{992} & {\cellcolor{gray!30}21.9} & 1471 & - & 886 & 995 & 793 & - & 993 & \bf{70} & - & 665 & {\cellcolor{gray!30}988} & 67.9 & - \\ 
 johnson32-2-4 & {\cellcolor{gray!30}\bf{465}} & \bf{481} & {\cellcolor{gray!30}6.9} & 30 & - & \bf{465} & \bf{481} & \bf{15} & - & \bf{481} & \bf{15} & - & 348 & {\cellcolor{gray!30}480} & 56.9 & MEM \\ 
 keller5 & {\cellcolor{gray!30}634} & \bf{755} & {\cellcolor{gray!30}33.0} & 10738 & - & \bf{645} & 757 & \bf{727} & - & 756 & 14455 & - & 587 & {\cellcolor{gray!30}749} & 44.9 & - \\ 
 p-hat1000-2 & {\cellcolor{gray!30}\bf{882}} & 974 & {\cellcolor{gray!30}19.4} & 2194 & - & 877 & 978 & \bf{749} & - & \bf{972} & 1603 & - & 769 & {\cellcolor{gray!30}955} & 40.9 & - \\ 
 p-hat1000-3 & {\cellcolor{gray!30}\bf{781}} & \bf{961} & {\cellcolor{gray!30}39.0} & 494 & - & 767 & 972 & 302 & - & 967 & \bf{102} & - & 698 & {\cellcolor{gray!30}938} & 54.8 & - \\ 
 p-hat500-3 & {\cellcolor{gray!30}\bf{400}} & 465 & {\cellcolor{gray!30}30.1} & 4577 & - & 392 & 471 & \bf{833} & - & \bf{464} & 5918 & - & 375 & {\cellcolor{gray!30}452} & 37.9 & - \\ 
 p-hat700-2 & {\cellcolor{gray!30}\bf{620}} & \bf{666} & {\cellcolor{gray!30}14.6} & 3741 & - & 614 & 676 & \bf{898} & - & 671 & 4521 & - & 601 & {\cellcolor{gray!30}656} & 18.0 & - \\ 
 p-hat700-3 & {\cellcolor{gray!30}\bf{552}} & 664 & {\cellcolor{gray!30}35.7} & 893 & - & 543 & 663 & \bf{568} & - & \bf{662} & 853 & - & 515 & {\cellcolor{gray!30}639} & 42.9 & - \\ 
 san1000 & {\cellcolor{gray!30}924} & \bf{991} & {\cellcolor{gray!30}13.6} & 91 & - & \bf{932} & \bf{991} & 87 & - & \bf{991} & \bf{27} & - & 730 & {\cellcolor{gray!30}990} & 53.1 & MEM \\ 
 san200-0.7-2 & {\cellcolor{gray!30}\bf{182}} & {\cellcolor{gray!30}\bf{182}} & {\cellcolor{gray!30}0} & 11190 & {\cellcolor{gray!30}\bf{1084.9}} & 177 & 188 & \bf{1642} & - & \bf{182} & 11724 & 1621.8 & 161 & 184 & 27.4 & MEM \\ 
 san400-0.5-1 & {\cellcolor{gray!30}\bf{386}} & {\cellcolor{gray!30}\bf{387}} & {\cellcolor{gray!30}0.5} & 10110 & - & 382 & 392 & \bf{1831} & - & \bf{387} & 9629 & - & 336 & 388 & 27.7 & - \\ 
 san400-0.7-1 & {\cellcolor{gray!30}\bf{305}} & \bf{379} & {\cellcolor{gray!30}41.3} & 15164 & - & 300 & \bf{379} & \bf{947} & - & \bf{379} & 16308 & - & 292 & {\cellcolor{gray!30}361} & 42.9 & - \\ 
 san400-0.7-2 & {\cellcolor{gray!30}\bf{321}} & \bf{383} & {\cellcolor{gray!30}33.9} & 13956 & - & 315 & 384 & \bf{853} & - & \bf{383} & 14084 & - & 303 & {\cellcolor{gray!30}371} & 39.8 & MEM \\ 
 san400-0.7-3 & {\cellcolor{gray!30}\bf{347}} & \bf{386} & {\cellcolor{gray!30}21.0} & 13287 & - & 339 & 387 & \bf{927} & - & \bf{386} & 11431 & - & 303 & {\cellcolor{gray!30}383} & 43.7 & MEM \\ 
 san400-0.9-1 & 249 & \bf{348} & 66.9 & 18649 & - & \bf{252} & 351 & \bf{544} & - & \bf{348} & 19184 & - & {\cellcolor{gray!30}278} & {\cellcolor{gray!30}300} & {\cellcolor{gray!30}22.0} & - \\ 

 sanr400-0.7 & {\cellcolor{gray!30}\bf{351}} & \bf{382} & 17.0 & 1954 & - & 349 & \bf{382} & \bf{978} & - & \bf{382} & 2011 & - & 349 & {\cellcolor{gray!30}379} & {\cellcolor{gray!30}16.8} & - \\ 
 
\hline
\end{tabular}
\caption{Comparison of HS-lub, HS-lb, HS-ub and hbfs in Maxclique instances. Instances at the top are mainly those that could not be solved by hbfs in their original paper (discrepancies are due to different hardware specifications). All algorithms are executed with 16 cpu-cores. Bodface is used to emphasize winning values when comparing HS-lub with respect to HS-lb and HS-ub. Grey back-ground is used to emphasize winning values when comparing HS-lub with respect to hbfs. An "-" indicates time-limit exceeded. A "MEM" indicates that the algorithm runs out of memory during the execution.}\label{table-maxclique}
\end{table*}

\end{landscape}

\begin{landscape}
\begin{table*}[th]
\scriptsize
\begin{tabular}{l|rrrrr|rrrr|rrr|rrrr}
\hline
 & \multicolumn{5}{c|}{HS-lub} & \multicolumn{4}{c|}{HS-lb} &  \multicolumn{3}{c|}{HS-ub} &  \multicolumn{4}{c}{hbfs-16} \\  
Inst. & lb & ub &  gap  &cores& t & lb & ub & cores & t & ub & cores & t & lb & ub & gap & t \\ \hline
 MANN-a27 & {\cellcolor{gray!30}\bf{252}} & {\cellcolor{gray!30}\bf{252}} & {\cellcolor{gray!30}0} & 691 & \bf{38.5} & \bf{252} & \bf{252} & \bf{557} & 128.3 & \bf{252} & 668 & 57.7 & {\cellcolor{gray!30}252} & {\cellcolor{gray!30}252} & {\cellcolor{gray!30}0} & {\cellcolor{gray!30}0.1} \\ 
 MANN-a45 & {\cellcolor{gray!30}\bf{690}} & {\cellcolor{gray!30}\bf{690}} & {\cellcolor{gray!30}0} & 1298 & {\cellcolor{gray!30}\bf{71.8}} & \bf{690} & \bf{690} & \bf{827} & 366.8 & \bf{690} & 1297 & 100.2 & {\cellcolor{gray!30}690} & {\cellcolor{gray!30}690} & {\cellcolor{gray!30}0} & 104.6 \\ 
 brock200-1 & \bf{165} & \bf{181} & 19.8 & 13064 & - & 163 & 184 & \bf{1471} & - & 182 & 11222 & - & {\cellcolor{gray!30}179} & {\cellcolor{gray!30}179} & {\cellcolor{gray!30}0} & {\cellcolor{gray!30}20.1} \\ 
 brock200-2 & {\cellcolor{gray!30}\bf{188}} & {\cellcolor{gray!30}\bf{188}} & {\cellcolor{gray!30}0} & \bf{918} & \bf{52.7} & \bf{188} & \bf{188} & 1368 & 332.2 & \bf{188} & \bf{918} & 70.9 & {\cellcolor{gray!30}188} & {\cellcolor{gray!30}188} & {\cellcolor{gray!30}0} & {\cellcolor{gray!30}0.6} \\ 
 brock200-3 & \bf{182} & 187 & 5.7 & 10545 & - & 181 & 187 & \bf{2112} & - & \bf{186} & 9920 & - & {\cellcolor{gray!30}185} & {\cellcolor{gray!30}185} & {\cellcolor{gray!30}0} & {\cellcolor{gray!30}1.5} \\ 
 brock200-4 & \bf{177} & \bf{184} & 8.2 & 10593 & - & 176 & 186 & \bf{2022} & - & 185 & 10082 & - & {\cellcolor{gray!30}183} & {\cellcolor{gray!30}183} & {\cellcolor{gray!30}0} & {\cellcolor{gray!30}3.6} \\ 
 c-fat200-1 & {\cellcolor{gray!30}\bf{188}} & {\cellcolor{gray!30}\bf{188}} & {\cellcolor{gray!30}0} & 36 & \bf{0.7} & \bf{188} & \bf{188} & \bf{24} & 0.9 & \bf{188} & 25 & 0.8 & {\cellcolor{gray!30}188} & {\cellcolor{gray!30}188} & {\cellcolor{gray!30}0} & {\cellcolor{gray!30}0.5} \\ 
 c-fat200-2 & {\cellcolor{gray!30}\bf{176}} & {\cellcolor{gray!30}\bf{176}} & {\cellcolor{gray!30}0} & 107 & \bf{1.0} & \bf{176} & \bf{176} & 93 & 3.2 & \bf{176} & \bf{89} & 1.6 & {\cellcolor{gray!30}176} & {\cellcolor{gray!30}176} & {\cellcolor{gray!30}0} & {\cellcolor{gray!30}0.6} \\ 
 c-fat200-5 & {\cellcolor{gray!30}\bf{142}} & {\cellcolor{gray!30}\bf{142}} & {\cellcolor{gray!30}0} & 289 & \bf{2.5} & \bf{142} & \bf{142} & \bf{224} & 10.6 & \bf{142} & 273 & 4.7 & {\cellcolor{gray!30}142} & {\cellcolor{gray!30}142} & {\cellcolor{gray!30}0} & {\cellcolor{gray!30}0.4} \\ 
 c-fat500-1 & {\cellcolor{gray!30}\bf{486}} & {\cellcolor{gray!30}\bf{486}} & {\cellcolor{gray!30}0} & 60 & \bf{13.0} & \bf{486} & \bf{486} & 49 & 18.9 & \bf{486} & \bf{47} & 13.6 & {\cellcolor{gray!30}486} & {\cellcolor{gray!30}486} & {\cellcolor{gray!30}0} & {\cellcolor{gray!30}11.3} \\ 
 c-fat500-10 & {\cellcolor{gray!30}\bf{374}} & {\cellcolor{gray!30}\bf{374}} & {\cellcolor{gray!30}0} & \bf{932} & \bf{31.3} & \bf{374} & \bf{374} & 1030 & 145.7 & \bf{374} & 952 & 40.5 & {\cellcolor{gray!30}374} & {\cellcolor{gray!30}374} & {\cellcolor{gray!30}0} & {\cellcolor{gray!30}6.3} \\ 
 c-fat500-2 & {\cellcolor{gray!30}\bf{474}} & {\cellcolor{gray!30}\bf{474}} & {\cellcolor{gray!30}0} & 179 & \bf{15.7} & \bf{474} & \bf{474} & \bf{152} & 21.8 & \bf{474} & 171 & 20.2 & {\cellcolor{gray!30}474} & {\cellcolor{gray!30}474} & {\cellcolor{gray!30}0} & {\cellcolor{gray!30}10.3} \\ 
 c-fat500-5 & {\cellcolor{gray!30}\bf{436}} & {\cellcolor{gray!30}\bf{436}} & {\cellcolor{gray!30}0} & \bf{450} & \bf{14.2} & \bf{436} & \bf{436} & 503 & 76.1 & \bf{436} & 464 & 26.6 & {\cellcolor{gray!30}436} & {\cellcolor{gray!30}436} & {\cellcolor{gray!30}0} & {\cellcolor{gray!30}8.8} \\ 
 hamming6-4 & {\cellcolor{gray!30}\bf{60}} & {\cellcolor{gray!30}\bf{60}} & {\cellcolor{gray!30}0} & 70 & \bf{0.3} & \bf{6m0} & \bf{60} & \bf{61} & 1.4 & \bf{60} & 62 & 0.6 & {\cellcolor{gray!30}60} & {\cellcolor{gray!30}60} & {\cellcolor{gray!30}0} & {\cellcolor{gray!30}0} \\ 
 hamming8-4 & \bf{238} & {\cellcolor{gray!30}\bf{240}} & 1.8 & 10656 & - & 236 & \bf{240} & \bf{2024} & - & \bf{240} & 10537 & - & {\cellcolor{gray!30}240} & {\cellcolor{gray!30}240} & {\cellcolor{gray!30}0} & {\cellcolor{gray!30}7.3} \\ 
 johnson16-2-4 & {\cellcolor{gray!30}\bf{112}} & {\cellcolor{gray!30}\bf{112}} & {\cellcolor{gray!30}0} & 327 & \bf{6.5} & \bf{112} & \bf{112} & \bf{311} & 22.7 & \bf{112} & \bf{311} & 12.4 & {\cellcolor{gray!30}112} & {\cellcolor{gray!30}112} & {\cellcolor{gray!30}0} & {\cellcolor{gray!30}6.3} \\ 
 johnson8-4-4 & {\cellcolor{gray!30}\bf{56}} & {\cellcolor{gray!30}\bf{56}} & {\cellcolor{gray!30}0} & 236 & \bf{2.1} & \bf{56} & \bf{56} & 250 & 12.1 & \bf{56} & \bf{227} & 2.7 & {\cellcolor{gray!30}56} & {\cellcolor{gray!30}56}  & {\cellcolor{gray!30}0} & {\cellcolor{gray!30}0.1} \\ 
 keller4 & \bf{158} & {\cellcolor{gray!30}\bf{160}} & 2.7 & 12087 & - & 156 & 161 & \bf{2044} & - & \bf{160} & 11551 & - & {\cellcolor{gray!30}160} & {\cellcolor{gray!30}160} & {\cellcolor{gray!30}0} & {\cellcolor{gray!30}1.8} \\ 
 p-hat1000-1 & \bf{973} & \bf{991} & 3.7 & 127 & - & 965 & \bf{991} & 22 & - & \bf{991} & \bf{20} & - & {\cellcolor{gray!30}990} & {\cellcolor{gray!30}990} & {\cellcolor{gray!30}0} & {\cellcolor{gray!30}215.7} \\ 
 p-hat300-1 & {\cellcolor{gray!30}\bf{292}} & {\cellcolor{gray!30}\bf{292}} & {\cellcolor{gray!30}0} & 315 & \bf{18.8} & \bf{292} & \bf{292} & 314 & 49.4 & \bf{292} & \bf{299} & 26.6 & {\cellcolor{gray!30}292} & {\cellcolor{gray!30}292} & {\cellcolor{gray!30}0} & {\cellcolor{gray!30}2.0} \\ 
 p-hat300-2 & \bf{268} & \bf{278} & 7.8 & 11512 & - & 266 & 281 & \bf{1441} & - & 279 & 11200 & - & {\cellcolor{gray!30}275} & {\cellcolor{gray!30}275} & {\cellcolor{gray!30}0} & {\cellcolor{gray!30}3.2} \\ 
 p-hat300-3 & \bf{242} & \bf{270} & 23.3 & 14852 & - & 235 & 279 & \bf{927} & - & \bf{270} & 14371 & - & {\cellcolor{gray!30}264} & {\cellcolor{gray!30}264} & {\cellcolor{gray!30}0} & {\cellcolor{gray!30}253.9} \\ 
 p-hat500-1 & {\cellcolor{gray!30}\bf{491}} & {\cellcolor{gray!30}\bf{491}} & {\cellcolor{gray!30}0} & 1033 & \bf{306.4} & \bf{491} & \bf{491} & 1028 & 718.7 & \bf{491} & \bf{970} & 716.4 & {\cellcolor{gray!30}491} & {\cellcolor{gray!30}491} & {\cellcolor{gray!30}0} & {\cellcolor{gray!30}9.4} \\ 
 p-hat500-2 & \bf{447} & \bf{472} & 11.3 & 7599 & - & 441 & 477 & \bf{1078} & - & \bf{472} & 5957 & - & {\cellcolor{gray!30}464} & {\cellcolor{gray!30}464} & {\cellcolor{gray!30}0} & {\cellcolor{gray!30}364.3} \\ 
 p-hat700-1 & \bf{687} & \bf{690} & 0.9 & 553 & - & 685 & 691 & \bf{215} & - & 691 & 406 & - & {\cellcolor{gray!30}689} & {\cellcolor{gray!30}689} & {\cellcolor{gray!30}0} & {\cellcolor{gray!30}32.1} \\ 
 san200-0.7-1 & \bf{162} & \bf{184} & 26.2 & 12170 & - & 160 & 185 & \bf{1236} & - & \bf{184} & 11583 & - & {\cellcolor{gray!30}170} & {\cellcolor{gray!30}170} & {\cellcolor{gray!30}0} & {\cellcolor{gray!30}41.5} \\ 
 san200-0.9-1 & {\cellcolor{gray!30}\bf{130}} & {\cellcolor{gray!30}\bf{130}} & {\cellcolor{gray!30}0} & 484 & \bf{1.7} & \bf{130} & \bf{130} & \bf{242} & 6.8 & \bf{130} & 4251 & 136.4 & {\cellcolor{gray!30}130} & {\cellcolor{gray!30}130} & {\cellcolor{gray!30}0} & {\cellcolor{gray!30}0.1} \\ 
 san200-0.9-2 & {\cellcolor{gray!30}\bf{140}} & {\cellcolor{gray!30}\bf{140}} & {\cellcolor{gray!30}0} & 6160 & \bf{193.7} & 138 & 174 & \bf{907} & - & 164 & 16004 & - & {\cellcolor{gray!30}140} & {\cellcolor{gray!30}140} & {\cellcolor{gray!30}0} & {\cellcolor{gray!30}0.8} \\ 
 san200-0.9-3 & \bf{142} & \bf{168} & 38.2 & 15339 & - & 139 & 173 & \bf{999} & - & \bf{168} & 13215 & - & {\cellcolor{gray!30}156} & {\cellcolor{gray!30}156} & {\cellcolor{gray!30}0} & {\cellcolor{gray!30}152.0} \\ 
 sanr200-0.7 & \bf{174} & 184 & 11.9 & 10737 & - & 173 & 184 & \bf{1937} & - & \bf{183} & 11048 & - & {\cellcolor{gray!30}182} & {\cellcolor{gray!30}182} & {\cellcolor{gray!30}0} & {\cellcolor{gray!30}7.0} \\ 
 sanr200-0.9 & \bf{141} & 167 & 38.2 & 14307 & - & 138 & 172 & \bf{1108} & - & \bf{166} & 12675 & - & {\cellcolor{gray!30}158} & {\cellcolor{gray!30}158} & {\cellcolor{gray!30}0} & {\cellcolor{gray!30}290.6} \\ 
 sanr400-0.5 & \bf{380} & \bf{388} & 4.3 & 1905 & - & 379 & \bf{388} & \bf{875} & - & \bf{388} & 1167 & - & {\cellcolor{gray!30}387} & {\cellcolor{gray!30}387} & {\cellcolor{gray!30}0} & {\cellcolor{gray!30}57.5} \\ 
\hline
\end{tabular}
\caption{Comparison of HS-lub, HS-lb, HS-ub and hbfs in Maxclique instances. Instances at the top are mainly those that could not be solved by hbfs in their original paper (discrepancies are due to different hardware specifications). All algorithms are executed with 16 cpu-cores. Bodface is used to emphasize winning values when comparing HS-lub with respect to HS-lb and HS-ub. Grey back-ground is used to emphasize winning values when comparing HS-lub with respect to hbfs. An "-" indicates time-limit exceeded. A "MEM" indicates that the algorithm runs out of memory during the execution.}\label{table-maxclique2}
\end{table*}
\end{landscape}

The first goal of our empirical evaluation is to study if, given a fixed number of cpu-cores ($16$ in our case), there is a dominant approach for HS-lub on how to use them.
Since we implement the hitting-vector component using CPLEX (which is multithreaded), we first investigate the impact of different ways to balance the load among the $16$ available cpu-cores on HS-lub.
We denote by HS-lub($x,y$) (with $x+y=16$) an execution where the lower and the upper bound loops are given $x$ and $y$ cores, respectively. 

In the first experiment, we solved all the instances with HS-lub(1,15), HS-lub(4,12), HS-lub(8,8), HS-lub(12,4) and HS-lub(15,1). We do not include these results for space reasons, but we observe that, except for a few cases in which the extreme work-load distribution was not effective, there was not a big difference among the different alternatives. Although this may seem unexpected, a similar observation is made in~\cite{DBLP:conf/cp/hbfs-multithread} when analyzing different load balances in CPLEX.

For the rest of the experiments we will restrict ourselves to HS-lub(8,8) which, in the following, we denote as HS-lub for short. Results regarding SPOT5, Pedigree and EHI are reported in Table~\ref{table-benchs} and results regarding Maxclique DIMACS are reported in Table~\ref{table-maxclique}. m


In our second experiment we compare HS-lub wrt HS-lb and HS-ub. In the tables we use boldface to emphasize best values with respect to this comparison.
It can be clearly seen that in the four benchmarks HS-lub is consistently superior than both HS-lb and HS-ub (see the domination of boldface values in HS-lub columns). In the following analysis we distinguish between SPOT5, Maxclique and pedigree instances, where there are instances that can and cannot be solved within the time limit, and EHI instances, where all instances are solved within the time limit.

Regarding SPOT5, Maxclique and pedigree instances: First, all instances solved by either HS-lb or HS-ub within the time limit ($45$ in total) are also solved by HS-lub. Moreover, HS-lub is always faster. The average speed-up of HS-lub wrt the best of HS-lb and HS-ub is of $1.85$ across all instances ($2.07$ for spot, $1.86$ for pedigree and $1.71$ for maxclique instances).
Second, HS-lub solves $2$ instances more than HS-lb (san200$\_$0.7$\_$2 and san200$\_$0.9$\_$2) and $2$ instances more than HS-ub (1504 and san200$\_$0.9$\_$2). Third, for those instances that none of the HS algorithms solve ($47$ in total), the lower bound of HS-lub is always better than that of HS-lb (except for $5$ instances) while the upper bound is either equal of better than the upper bound of HS-ub in most of the instances (only in $12$ instances the upper bound is slightly worst). 

Regarding EHI instances: HS-lub slightly outperforms both HS-lb and HS-ub. For EHI-85 and EHI-90, the speed-up wrt HS-lb is of $1.2$ and $1.11$, respectively, while of $1.10$ and $1.10$ wrt HS-ub, respectively.

Next, we compare the average number of cores used by HS-lub wrt HS-lb (resp. HS-ub) on instances solved by HS-lb (resp. HS-ub). Note that, on SPOT5, Pedigree and Maxclique benchmarks, all instances solved by either HS-lb or HS-ub are also solved by HS-lub but HS-lb and HS-ub do not solve the entire same set of instances. Compared to HS-lb, HS-lub needs $531.64$ cores while HS-lb needs $347.66$. Compared to HS-ub, HS-lub needs $711.20$ cores while HS-ub needs $867.73$. The first thing to note is that HS-lb uses less cores but, since HS-lub is faster than HS-lb, those cores are more time-consuming to obtain (i.e., HS-lub benefits from the less time consuming cost-bounded cores). The second thing to note is that HS-ub uses more cores but, although those cores are less time consuming to obtain, HS-lub benefits from minimum cost cores. For EHI instances, where all instances are solved by all approaches, HS-lub needs $17.64$ cores while HS-lb and HS-ub need $13.34$ and $12.02$, respectively. We find the fact that HS-ub needs less cores than HS-lb  counter-intuitive and we still have not found and explanation for it. As a consequence, in this benchmark HS-lub needs more cores than either HS-lb or HS-ub, but still its performance is better.

In our third and last experiment, we analyze the anytime nature of HS-lub and compare it with the parallel hbfs implementation of Toulbar2 (using the $16$ available cpu-cores and noted hbfs-16). In the tables we use grey background to emphasize best values with respect to this comparison. 

\begin{figure}[t]
\input{figuraSupl1}
\caption{Selected executions of HS-lub and hbfs-16 on instances from SPOT5 (1506 and 414), and Maxclique (brock400-1 and p-hat300-1) benchmarks. Plots show the evolution of the upper and lower bounds as a function of time.}
\label{fig:anytime}
\end{figure}
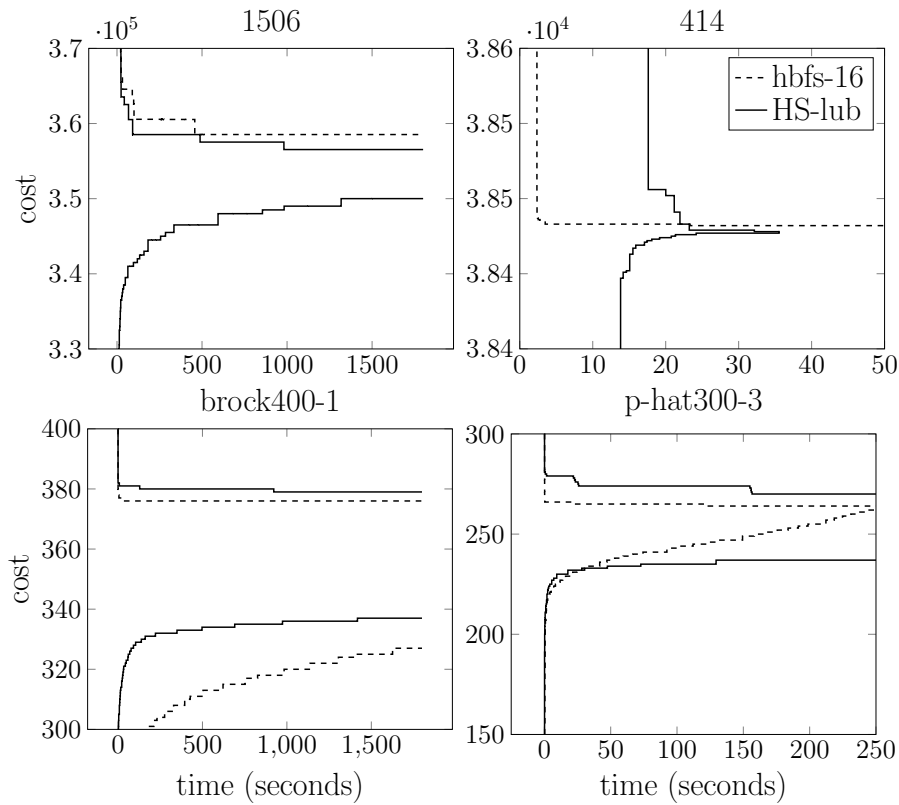

In the SPOT5 benchmark, HS-lub is consistently better than hbfs-16. First, HS-lub solves in a few seconds $8$ instances where hbfs-16 times out.
Second, in the $7$ instances where both algorithms time out, HS-lub always obtains significantly better lower bounds while superior upper bounds in $4$ of them.
The optimality gap is highly reduced from an average of $90.17$ (hbfs-16) to an average of $9.0$ (HS-lub). Third, in the $3$ instances solved by both algorithms, HS-lub is always faster (and up to $3$ orders of magnitude in instance $503$). Figure~\ref{fig:anytime} gives further detail of the anytime behaviour of both algorithms on instances 1506 and 414. On instance 1506, which is solved only by HS-lub, both upper and lower bounds of HS-lub are better along time. However, a more usual behaviour is the one showed on instance 414 where hbfs-16 improves the upper bound very fast and then further improvements are slow and hard. On the other hand, HS-lub improves its lower bound faster than its upper bound, but it is able to keep improving its upper bound progressively. 

We obtain similar results for pedigree instances. In particular, HS-lub solves $2$ more instances than hbfs-16 in $2.3$ and $526.0$ seconds, respectively; none of the algorithms solve one instance (pedigree40) but HS-lub obtains better bounds reducing the optimality gap from $77.9$ to $2.9$; for those instances solved by both algorithms, the solving time is either similar or HS-lub is up-to two orders of magnitude better (see pedigree25 and pedigree44 instances). The anytime behaviour of both algorithms is the same as in the previous benchmark.



For the Maxclique instances the dominance of HS-lub is not so clear. We distinguish between instances solved very easily by hbfs-16 ($32$ instances) and instances where hbfs-16 either times out ($20$ instances) or runs out of memory ($5$ instances). In the first case, hbfs-16 clearly outperforms HS-lub. In particular, HS-lub does not solve $15$ of those instances while for the other $17$ instances, the time is usually greater. In the second case, hbfs-16 obtains in general better upper bounds, while HS-lub obtains better lower bounds. 
However, the optimality gap of HS-lub is better than that of hbfs-16 (except for $6$ instances). This improvement ranges from $3\%$ up to $46\%$, which means that HS-lub compensates the inferiority in the upper bound by the superiority in the lower bound. Hbfs-16 runs out of memory in $5$ instances, while HS-lub does not which indicates that the HS approach is less memory demanding. Moreover, HS-lub solves $1$ instance where hbfs-16 times out (san200$\_$0.7$\_$2 instance). Figure~\ref{fig:anytime} gives further detail of the anytime behaviour of both algorithms on instances brock400-1 and p-hat300-3. brock400-1 exemplifies the behaviour of unsolved instances. Both algorithms improve the upper bound very fast at the beginning of the execution until it stabilizes to a given value, hbfs-16 being better than HS-lub. The lower bound improves steadily along time, HS-lub being better than hbfs-16. p-hat300-3 exemplifies the behaviour of instances easily solved by hbfs-16. Both algorithms improve the bounds relatively fast at the very beginning, but only hbfs-16 is able to keep improving its lower bound.



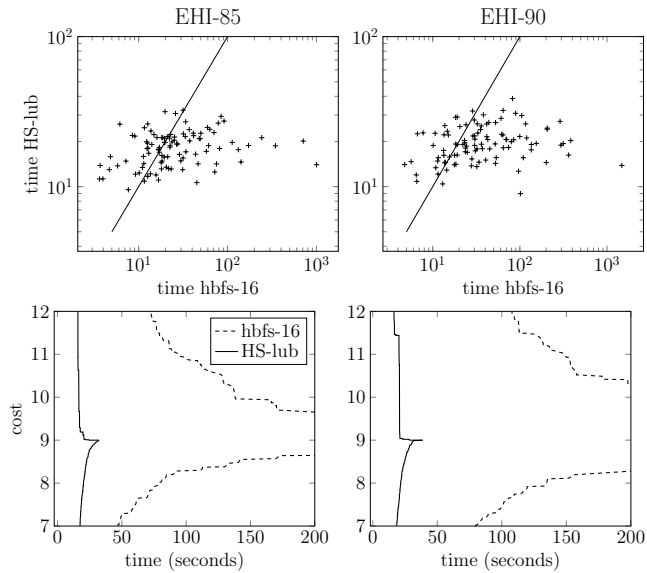
\begin{figure}[t]
\input{figuraSupl2}
\caption{Results on EHI-85 (left column) and EHI-90 (right column) benchmarks. Plots on the first row (note the log scale) show, for each instance (each dot), solving time of HS-lub wrt hbfs-16. Black line indicates equal solving times. Plots on the second row show average lower and upper bounds of HS-lub and hbfs-16 as a function of time. Each benchmark has $100$ instances. The average is meaningful because the optimum is $9$ for all of them.}\label{fig-ehi}
\end{figure}

For EHI instances (see Table~\ref{table-benchs}), HS-lub outperforms hbfs-16: on average, the speed-up of HS-lub is of $2.88$ and $3.74$ on EHI-85 and EHI-90, respectively. A number of instances are relatively hard for hbfs-16. In particular, solving times for $4$ and $10$ instances in EHI-85 and EHI-90, respectively, are greater than $200$ seconds. Even discarding those outliers, HS-lub generally outperforms hbfs-16: on the $58.3\%$ and $70.3\%$ (EHI-85 and EHI-90, respectively) of those non-outliers instances HS-lub is faster than hbfs-16. Figure~\ref{fig-ehi} shows solving times of HS-lub wrt hbfs-16 for all instances, and the average upper and lower bounds for both algorithms as a function of time. It confirms that HS-lub is more convenient than hbfs-16 in this benchmark.





\section{Conclusion and Future Work}
We have presented the first parallel anytime algorithm for discrete optimization based on the implicit hitting set approach. Our multithread implementation is simpler than other multithread alternatives such as hbfs. Its work-load is well balanced and the overhead of the communication is very low. 

We have demonstrated the potential of our approach in several standard benchmarks of  the WCSP problem where we have shown that it often outperforms the state-of-the-art alternative parallel version of Toulbar2. We believe that these results show that our approach has potential.

There is much room for improvement in our implementation. We plan to incorporate techniques that have boosted performance in MaxSAT and Pseudo-Boolean Optimization such as reduced cost fixing, weight-aware cost extraction \cite{DBLP:conf/sat/IHS-PB2} or greedy hitting vectors \cite{DBLP:phd/ca/Davies14,DBLP:conf/cp/DelisleB13}. 
Our experience is that current implementations of the HS approach for WCSP are not effective for WCSPs with large domain variables. In our experiments, the explanation was the bad performance of SAT solvers, which made extremely inefficient to obtain maximal cores. To overcome this limitation we also plan to explore in the future alternative SAT encodings more suitable for that type of problems or switching from SAT solvers to CP solvers. Since the hitting vectors computed by HS-ub do not require to be optimal, the resulting problem is essentially a set of standard clauses plus one pseudo-boolean constraint, for which there are many efficient SAT encodings. Therefore, replacing the IP solver by a SAT solver may also be useful here.

\bibliography{mybibfile} 

\end{document}

%% file: figuraEmma1.tex
\footnotesize
    \centering  
\begin{tabular}{c}  
\begin{tabular}[b]{cc|c}
$x_1$ & $x_2$ & $f(x_1,x_2)$\\
\cline{1-3}
a & a & 0 \\
a & b & 20 \\
b & a & 5 \\
b & b & 20 \\
\multicolumn{3}{c}{}\\
$x_2$ & $x_3$ & $g(x_2,x_3)$\\
\cline{1-3}
a & a & 20\\
a & b & 20\\
b & a & 5\\
b & b & 0\\
\end{tabular}
\hspace{1cm}
\begin{tikzpicture}[scale=0.9]
    \draw [<->,very thick] (0,2.5) node (yaxis) [above] {}
        |- (2.5,0) node (xaxis) [right] {};

    \coordinate (x0) at (0,0);
    \coordinate (x1) at (0.5,0);
    \coordinate (x2) at (1,0);
    \coordinate (x3) at (1.5,0);
    \coordinate (x4) at (2,0);
    \coordinate (y1) at (0,0.5);
    \coordinate (y2) at (0,1);
    \coordinate (y3) at (0,1.5);
    \coordinate (y4) at (0,2);
    \draw node[below of = x0, yshift=5mm]{0};
    \draw node[below of = x1, yshift=5mm]{5};
    \draw node[below of = x4, yshift=5mm]{20};
    \draw node[left of = x0, xshift=5mm]{0};
    \draw node[left of = y1, xshift=5mm]{5};
    \draw node[left of = y4, xshift=5mm]{20};

    \draw [dashed, thin, gray] (0, 0.5) -- (2.4,0.5);
    \draw [dashed, thin, gray] (0, 2) -- (2.4,2);
    \draw [dashed, thin, gray] (0.5, 0) -- (0.5, 2.4);
    \draw [dashed, thin, gray] (2, 0) -- (2, 2.4);

    \fill[red] (0,0) circle (2pt);
    \fill[red] (0,0.5) circle (2pt);
    \fill[red] (0.5,0) circle (2pt);
    \fill[red] (0.5,0.5) circle (2pt);
    \fill[blue] (0,2) circle (2pt);
    \fill[blue] (0.5,2) circle (2pt);
    \fill[blue] (2,2) circle (2pt);
    \fill[blue] (2,0.5) circle (2pt);
    \fill[blue] (2,0) circle (2pt);
\end{tikzpicture}
\end{tabular}

%% file: figuraEmma2.tex
    
    \centering  
\begin{tikzpicture}[scale=1]
    \def\u{0.27}
    \coordinate (c) at (4*\u, 8*\u);
    
    \fill[gray!20] (0,0) rectangle ++(c);

    \fill[black!60!green] (8*\u, 0*\u) circle (3pt);  
    \draw [<->,very thick] (0,3) node (yaxis) [above] {}
        |- (3,0) node (xaxis) [right] {};
    \draw[step=\u cm, gray, very thin, dashed] (0,0) grid (2.9,2.9); 

    \coordinate (x0) at (0,0);
    \coordinate (x1) at (0.5,0);
    \coordinate (x2) at (1,0);
    \coordinate (x3) at (1.5,0);
    \coordinate (x4) at (2,0);
    \coordinate (x5) at (2.5,0);
    \coordinate (y1) at (0,0.5);
    \coordinate (y2) at (0,1);
    \coordinate (y3) at (0,1.5);
    \coordinate (y4) at (0,2);
    \coordinate (y5) at (0,2.5);

    \fill[red] (c) circle (2pt);
    \draw[dashed, red] (yaxis |- c) node[left] {}
        -| (xaxis -| c) node[below] {};
    

    \draw[red] (5*\u,0*\u) circle (2pt) node (h) {} 
    node[black, above, font=\boldmath, xshift=-2mm, yshift=1mm] {\large $\vec{h}$};

   \fill[red] (7*\u, 3*\u) circle (1.5pt) node (k1) {}
   node[black, above, font=\boldmath, xshift=-3mm, yshift=0mm]
   {\large $\vec{k}$};

    \draw[->, thick] (h) -- (k1);

    \fill[blue] (8*\u, 3*\u) circle (1.5pt);
    \fill[blue] (7*\u, 4*\u) circle (1.5pt);

    \draw[red] (1*\u,9*\u) circle (2pt) node (h) {} node[black, above, font=\boldmath, xshift=-1mm,yshift=1mm] {\large $\vec{h'}$};
    \fill[red] (3*\u, 9*\u) circle (1.5pt) node (k1) {} node[black, above, font=\boldmath, xshift=3mm,yshift=1mm] {\large $\vec{k'}$};
    \draw[->, thick] (h) -- (k1);

    \fill[blue] (3*\u, 10*\u) circle (1.5pt);
    \fill[blue] (4*\u, 9*\u) circle (1.5pt);
    \fill[blue] (5*\u, 8*\u) circle (1.5pt);

\end{tikzpicture}

%% file: figuraSupl1.tex
\centering
\scalebox{0.7}{
\begin{tikzpicture}
\begin{axis}[
    title={1506},
    ylabel={cost},
    ymax=370000,
    ymin=330000,
    legend pos=north east,
    legend cell align=left,
    label style={font=\LARGE},
    x tick label style={
        font=\Large,
        /pgf/number format/.cd,
        set thousands separator={}
    },
    y tick label style={font=\Large},
    title style={font=\LARGE}
]

\addplot [black, dashed, thick] table [header=false, x expr={\thisrowno{1}}, y expr={(\thisrowno{0}==-1?\thisrowno{2}:NaN)}] {./dat/1506_VAC_toulbar_mpi_n16.dat}; 
\addplot  [black, dashed, thick] table [header=false, x expr={\thisrowno{1}}, y expr={(\thisrowno{0}==-2?\thisrowno{2}:NaN)}] {./dat/1506_VAC_toulbar_mpi_n16.dat}; 

\addplot [black, thick] table [header=false, x expr={\thisrowno{1}}, y expr={(\thisrowno{0}==-1?\thisrowno{2}:NaN)}] {./dat/1506_VAC_ihs_merged_mt_a3_t4_m8.dat};
\addplot  [black, thick] table [header=false, x expr={\thisrowno{1}}, y expr={(\thisrowno{0}==-2?\thisrowno{2}:NaN)}] {./dat/1506_VAC_ihs_merged_mt_a3_t4_m8.dat};

\end{axis}
\end{tikzpicture}

\begin{tikzpicture}
\begin{axis}[
    restrict y to domain=30000:48000,
    title={414},
    ymax=38600,
    ymin=38400,
    xmax=50,    
    xmin=0,
    legend pos=north east,
    legend cell align=left,
    label style={font=\LARGE},
    tick label style={font=\Large},
    title style={font=\LARGE},
    legend style={font=\LARGE}
]

\addplot [black, dashed, thick] table [header=false, x expr={\thisrowno{1}}, y expr={\thisrowno{2}}] {./dat/414_VAC_toulbar_mpi_n16_1.dat}; 
\addplot  [black, dashed, thick] table [header=false, x expr={\thisrowno{1}}, y expr={\thisrowno{2}}] {./dat/414_VAC_toulbar_mpi_n16_2.dat}; 

\addplot [black, thick] table [header=false, x expr={\thisrowno{1}}, y expr={\thisrowno{2}}] {./dat/414_VAC_ihs_merged_mt_a3_t4_m8_1.dat};
\addplot  [black, thick] table [header=false, x expr={\thisrowno{1}}, y expr={\thisrowno{2}}] {./dat/414_VAC_ihs_merged_mt_a3_t4_m8_2.dat};



  \legend{hbfs-16, , HS-lub, }  
\end{axis}
\end{tikzpicture}
}

\scalebox{0.7}{
\begin{tikzpicture}
\begin{axis}[
    title={brock400-1},
    xlabel={time (seconds)},
    ylabel={cost},
    ymax=400,
    ymin=300,
    legend pos=north east,
    legend cell align=left,
    label style={font=\LARGE},
    tick label style={font=\Large},
    title style={font=\LARGE},
    legend style={font=\Large}
]

\addplot [black, dashed, thick] table [header=false, x expr={\thisrowno{1}}, y expr={(\thisrowno{0}==-1?\thisrowno{2}:NaN)}] {./dat/brock400_1.clq_VAC_toulbar_mpi_n16.dat}; 
\addplot  [black, dashed, thick] table [header=false, x expr={\thisrowno{1}}, y expr={(\thisrowno{0}==-2?\thisrowno{2}:NaN)}] {./dat/brock400_1.clq_VAC_toulbar_mpi_n16.dat}; 

\addplot [black, thick] table [header=false, x expr={\thisrowno{1}}, y expr={(\thisrowno{0}==-1?\thisrowno{2}:NaN)}] {./dat/brock400_1.clq_VAC_ihs_merged_mt_a3_t4_m8.dat};
\addplot  [black, thick] table [header=false, x expr={\thisrowno{1}}, y expr={(\thisrowno{0}==-2?\thisrowno{2}:NaN)}] {./dat/brock400_1.clq_VAC_ihs_merged_mt_a3_t4_m8.dat};

\end{axis}
\end{tikzpicture}
\begin{tikzpicture}
\begin{axis}[
    title={p-hat300-3},
    xlabel={time (seconds)},
    ymax=300,
    ymin=150,
    xmax=250,    
    legend pos=north east,
    legend cell align=left,
    label style={font=\LARGE},
    tick label style={font=\Large, 
        /pgf/number format/.cd,
        1000 sep={},
        precision=0
    },
    title style={font=\LARGE},
    legend style={font=\Large}
]

\addplot [black, dashed, thick] table [header=false, x expr={\thisrowno{1}}, y expr={(\thisrowno{0}==-1?\thisrowno{2}:NaN)}] {./dat/p_hat300-3.clq_VAC_toulbar_mpi_n16.dat}; 
\addplot  [black, dashed, thick] table [header=false, x expr={\thisrowno{1}}, y expr={(\thisrowno{0}==-2?\thisrowno{2}:NaN)}] {./dat/p_hat300-3.clq_VAC_toulbar_mpi_n16.dat}; 

\addplot [black, thick] table [header=false, x expr={\thisrowno{1}}, y expr={(\thisrowno{0}==-1?\thisrowno{2}:NaN)}] {./dat/p_hat300-3.clq_VAC_ihs_merged_mt_a3_t4_m8.dat};
\addplot  [black, thick] table [header=false, x expr={\thisrowno{1}}, y expr={(\thisrowno{0}==-2?\thisrowno{2}:NaN)}] {./dat/p_hat300-3.clq_VAC_ihs_merged_mt_a3_t4_m8.dat};

\end{axis}
\end{tikzpicture}
}

%% file: figuraSupl2.tex
    \centering
\begin{tikzpicture}[scale=0.5] 
\begin{axis}[
    xmode=log,
    ymode=log,
    title={EHI-85},
    ylabel={time HS-lub},
    xlabel={time hbfs-16},
    ymax=100,
    label style={font=\Large},
    tick label style={
        font=\Large,
        /pgf/number format/.cd,
        1000 sep={}
    },
    title style={font=\LARGE}
]

\addplot [black] coordinates { (5,5) (10000,10000) };

\addplot [only marks, mark = +, black, ultra thin] table [header=false, x index=0, y index=1] {./dat/ehi-85-toulbar-a3.dat}; 


\end{axis}
\end{tikzpicture}
\begin{tikzpicture}[scale=0.5]
\begin{axis}[
    xmode=log,
    ymode=log,
    title={EHI-90},
    xlabel={time hbfs-16},
    ymax=100,
    legend pos=north east,
    legend cell align=left,
    label style={font=\Large},
    tick label style={
        font=\Large,
        /pgf/number format/.cd,
        1000 sep={}
    },
    title style={font=\LARGE}
]

\addplot [only marks, mark = +, black, ultra thin] table [header=false, x index=0, y index=1] {./dat/ehi-90-toulbar-a3.dat}; 
\addplot [black] coordinates { (5,5) (1000,1000) };

\end{axis}
\end{tikzpicture}

\begin{tikzpicture}[scale=0.5]  
\begin{axis}[
    restrict y to domain=7:13,
    xlabel={time (seconds)},
    ylabel={cost},
    ymax=12,
    ymin=7,
    xmax=200,
    legend pos=north east,
    legend cell align=left,
    label style={font=\Large},
    x tick label style={
        font=\Large,
        /pgf/number format/.cd,
        set thousands separator={}
    },
    y tick label style={font=\Large},
    title style={font=\LARGE},
    legend style={font=\Large}
]

\addplot [black, dashed, thick] table [header=false, x index=0, y index=1] {./dat/t_avg_toulbar_ub_ehi-85.dat}; 
\addplot  [black, dashed, thick] table [header=false, x index=0, y index=1] {./dat/t_avg_toulbar_lb_ehi-85.dat}; 

\addplot [black, thick] table [header=false, x index=0, y index=1] {./dat/t_avg_a3_ub_ehi-85.dat};
\addplot  [black, thick] table [header=false, x index=0, y index=1] {./dat/t_avg_a3_lb_ehi-85.dat};

  \legend{hbfs-16, , HS-lub, }  
\end{axis}
\end{tikzpicture}
\begin{tikzpicture}[scale=0.5]  
\begin{axis}[
    restrict y to domain=7:13,
    xlabel={time (seconds)},
    ymax=12,
    ymin=7,
    xmax=200,
    legend pos=north east,
    legend cell align=left,
    label style={font=\Large},
    tick label style={
        font=\Large,
        /pgf/number format/.cd,
        1000 sep={}
    },
    title style={font=\LARGE}
]

\addplot [black, dashed, thick] table [header=false, x index=0, y index=1] {./dat/t_avg_toulbar_ub_ehi-90.dat}; 
\addplot  [black, dashed, thick] table [header=false, x index=0, y index=1] {./dat/t_avg_toulbar_lb_ehi-90.dat}; 

\addplot [black, thick] table [header=false, x index=0, y index=1] {./dat/t_avg_a3_ub_ehi-90.dat};
\addplot  [black, thick] table [header=false, x index=0, y index=1] {./dat/t_avg_a3_lb_ehi-90.dat};

\end{axis}
\end{tikzpicture}

%% file: sample-sigconf.bbl
\begin{thebibliography}{10}

\bibitem{DBLP:conf/cp/hbfs}
David Allouche, Simon de~Givry, George Katsirelos, Thomas Schiex, and Matthias Zytnicki.
\newblock Anytime hybrid best-first search with tree decomposition for weighted {CSP}.
\newblock In {\em {CP} 2015, Cork, Ireland, 2015, Proceedings}, pages 12--29. Springer, 2015.

\bibitem{DBLP:journals/ai/AnsoteguiBL13}
Carlos Ans{\'{o}}tegui, Maria~Luisa Bonet, and Jordi Levy.
\newblock Sat-based maxsat algorithms.
\newblock {\em Artif. Intell.}, 196:77--105, 2013.

\bibitem{DBLP:conf/cp/hbfs-multithread}
Abdelkader Beldjilali, Pierre Montalbano, David Allouche, George Katsirelos, and Simon de~Givry.
\newblock Parallel hybrid best-first search.
\newblock In {\em {CP} 2022, July 31 to August 8, 2022, Haifa, Israel}, volume 235 of {\em LIPIcs}, pages 7:1--7:10, 2022.

\bibitem{DBLP:journals/constraints/BensanaLV99}
E.~Bensana, Michel Lema{\^{\i}}tre, and G{\'{e}}rard Verfaillie.
\newblock Earth observation satellite management.
\newblock {\em Constraints An Int. J.}, 4(3):293--299, 1999.

\bibitem{DBLP:conf/sat/BergBP20}
Jeremias Berg, Fahiem Bacchus, and Alex Poole.
\newblock Abstract cores in implicit hitting set maxsat solving.
\newblock In {\em {SAT} 2020, Alghero, Italy, 2020}, pages 277--294. Springer, 2020.

\bibitem{cadical}
Armin Biere, Katalin Fazekas, Mathias Fleury, and Maximillian Heisinger.
\newblock {CaDiCaL}, {Kissat}, {Paracooba}, {Plingeling} and {Treengeling} entering the {SAT Comp. 2020}.
\newblock In {\em Proc.~of {SAT Competition} 2020 -- Solver and Benchmark Descriptions}, pages 51--53, 2020.

\bibitem{DBLP:journals/constraints/CabonGLSW99}
Bertrand Cabon, Simon de~Givry, Lionel L., T.~Schiex, and J.~P. Warners.
\newblock Radio link frequency assignment.
\newblock {\em Constraints An Int. J.}, 4(1):79--89, 1999.

\bibitem{DBLP:conf/stacs/CooperGS20}
Martin~C. Cooper, Simon de~Givry, and Thomas Schiex.
\newblock Graphical models: Queries, complexity, algorithms.
\newblock In {\em {STACS} 2020,}, volume 154 of {\em LIPIcs}, pages 4:1--4:22, 2020.

\bibitem{cplex2009v12}
IBM~ILOG Cplex.
\newblock V12. 1: User’s manual for cplex.
\newblock {\em IBM Inc.}, 46(53), 2009.

\bibitem{DBLP:phd/ca/Davies14}
Jessica Davies.
\newblock {\em Solving {MAXSAT} by Decoupling Optimization and Satisfaction}.
\newblock PhD thesis, University of Toronto, Canada, 2014.

\bibitem{minibuckets}
Rina Dechter and Irina Rish.
\newblock Mini-buckets: {A} general scheme for bounded inference.
\newblock {\em J. {ACM}}, 50(2):107--153, 2003.

\bibitem{DBLP:conf/cp/DelisleB13}
Erin Delisle and Fahiem Bacchus.
\newblock Solving weighted csps by successive relaxations.
\newblock In {\em {CP} 2013, Uppsala, Sweden, September 16-20, 2013. Proceedings}, volume 8124 of {\em LNCS}, pages 273--281. Springer, 2013.

\bibitem{DBLP:books/fm/GareyJ79}
M.~R. Garey and David~S. Johnson.
\newblock {\em Computers and Intractability: {A} Guide to the Theory of NP-Completeness}.
\newblock W. H. Freeman, 1979.

\bibitem{DBLP:journals/tplp/GentMNMPMU18}
Ian~P. Gent, Ian Miguel, Peter Nightingale, Ciaran McCreesh, Patrick Prosser, Neil C.~A. Moore, and Chris Unsworth.
\newblock A review of literature on parallel constraint solving.
\newblock {\em Theory Pract. Log. Program.}, 18(5-6):725--758, 2018.

\bibitem{DBLP:reference/fai/GentPP06}
Ian~P. Gent, Karen~E. Petrie, and Jean{-}Fran{\c{c}}ois Puget.
\newblock Symmetry in constraint programming.
\newblock In Francesca Rossi, Peter van Beek, and Toby Walsh, editors, {\em Handbook of Constraint Programming}, volume~2 of {\em Foundations of AI}. Elsevier, 2006.

\bibitem{DBLP:journals/constraints/HurleyOAKSZG16}
B.~Hurley, B.~O'Sullivan, D.~Allouche, G.~Katsirelos, T.~Schiex, M.~Zytnicki, and S.~de~Givry.
\newblock Multi-language evaluation of exact solvers in graphical model discrete optimization.
\newblock {\em Constraints An Int. J.}, 21(3):413--434, 2016.

\bibitem{Larrosa24}
Javier Larrosa, Conrado Martinez, and Emma Rollon.
\newblock Theoretical and empirical analysis of cost-function merging for implicit hitting set wcsp solving.
\newblock In {\em AAAI 2024; Vancouver, Canada}. AAAI Press, 2024.

\bibitem{DBLP:books/sp/18/LynceMM18}
In{\^{e}}s Lynce, Vasco~M. Manquinho, and Ruben Martins.
\newblock Parallel maximum satisfiability.
\newblock In Youssef Hamadi and Lakhdar Sais, editors, {\em Handbook of Parallel Constraint Reasoning}, pages 61--99. Springer, 2018.

\bibitem{DBLP:journals/jair/MalapertRR16}
A.~Malapert, J.C. R{\'{e}}gin, and M.~Rezgui.
\newblock Embarrassingly parallel search in constr. programming.
\newblock {\em J. Artif. Intell. Res.}, 57:421--464, 2016.

\bibitem{DBLP:reference/fai/MeseguerRS06}
Pedro Meseguer, Francesca Rossi, and Thomas Schiex.
\newblock Soft constraints.
\newblock In Francesca Rossi, Peter van Beek, and Toby Walsh, editors, {\em Handbook of Constraint Programming}, volume~2 of {\em Foundations of AI}, pages 281--328. Elsevier, 2006.

\bibitem{DBLP:journals/constraints/MorgadoHLPM13}
Ant{\'{o}}nio Morgado, Federico Heras, Mark~H. Liffiton, Jordi Planes, and Jo{\~{a}}o Marques{-}Silva.
\newblock Iterative and core-guided maxsat solving: {A} survey and assessment.
\newblock {\em Constraints An Int. J.}, 18(4):478--534, 2013.

\bibitem{DBLP:books/daglib/nemhauser}
George~L. Nemhauser and Laurence~A. Wolsey.
\newblock {\em Integer and Combinatorial Optimization}.
\newblock Wiley interscience series in discrete mathematics and optimization. Wiley, 1988.
\newblock \href {https://doi.org/10.1002/9781118627372} {\path{doi:10.1002/9781118627372}}.

\bibitem{DBLP:journals/jair/OttenD17}
Lars Otten and Rina Dechter.
\newblock {AND/OR} branch-and-bound on a computational grid.
\newblock {\em J. Artif. Intell. Res.}, 59:351--435, 2017.

\bibitem{DBLP:conf/kr/SaikkoDAJ18}
Paul Saikko, Carmine Dodaro, Mario Alviano, and Matti J{\"{a}}rvisalo.
\newblock A hybrid approach to optimization in answer set programming.
\newblock In {\em {KR} 2018, Tempe, Arizona, 30 October - 2 November 2018}, pages 32--41. {AAAI} Press, 2018.

\bibitem{DBLP:conf/cp/IHS-PB1}
Pavel Smirnov, Jeremias Berg, and Matti J{\"{a}}rvisalo.
\newblock Pseudo-boolean optimization by implicit hitting sets.
\newblock In {\em {CP} 2021,}, volume 210 of {\em LIPIcs}, pages 51:1--51:20, 2021.

\bibitem{DBLP:conf/sat/IHS-PB2}
Pavel Smirnov, Jeremias Berg, and Matti J{\"{a}}rvisalo.
\newblock Improvements to the implicit hitting set approach to pseudo-boolean optimization.
\newblock In {\em {SAT} 2022, Haifa, Israel}, volume 236 of {\em LIPIcs}, pages 13:1--13:18, 2022.

\bibitem{DBLP:journals/bioinformatics/ViricelGSB18}
Cl{\'{e}}ment Viricel, Simon de~Givry, Thomas Schiex, and Sophie Barbe.
\newblock Cost function network-based design of protein-protein interactions: predicting changes in binding affinity.
\newblock {\em Bioinform.}, 34(15):2581--2589, 2018.

\bibitem{DBLP:journals/bioinformatics/VucinicSRBS20}
Jelena Vucinic, David Simoncini, Manon Ruffini, Sophie Barbe, and Thomas Schiex.
\newblock Positive multistate protein design.
\newblock {\em Bioinform.}, 36(1):122--130, 2020.

\bibitem{yeoh12}
William Yeoh and Makoto Yokoo.
\newblock Distributed problem solving.
\newblock {\em AI Magazine}, 33(3):53--65, 2012.

\end{thebibliography}
